\pdfoutput=1

\documentclass[11pt]{article}
\usepackage[dvipsnames]{xcolor}

\usepackage[final]{acl}

\usepackage{times}
\usepackage{latexsym}

\usepackage[T1]{fontenc}

\usepackage[utf8]{inputenc}

\usepackage{microtype}

\usepackage{inconsolata}
\usepackage{enumitem}
\usepackage{amsmath}
\usepackage{dsfont}
\usepackage{graphicx}
\usepackage{multicol}
\usepackage{multirow}
\usepackage{booktabs}
\usepackage{amssymb}
\usepackage{pifont}
\usepackage{tabularx}
\usepackage{wrapfig}
\newcommand{\cmark}{\ding{51}}%
\newcommand{\xmark}{\ding{55}}%
\DeclareMathOperator*{\argmax}{arg\,max}
\DeclareMathOperator*{\argmin}{arg\,min}
\newcommand*{\Scale}[2][4]{\scalebox{#1}{$#2$}}%
\newcommand{\Sref}[1]{\S\ref{#1}}

\title{Don't Hallucinate, Abstain:  \\Identifying LLM Knowledge Gaps via Multi-LLM Collaboration}

\author{Shangbin Feng\textsuperscript{1} \ \ \ \ \ \ \ Weijia Shi\textsuperscript{1} \ \ \ \ \ \ \ Yike Wang\textsuperscript{2} \\
\textbf{Wenxuan Ding\textsuperscript{3}} \ \ \ \textbf{Vidhisha Balachandran\textsuperscript{4}} \ \ \ \textbf{Yulia Tsvetkov\textsuperscript{1}} \\
\textsuperscript{1}University of Washington \ \ \ \ \ \textsuperscript{2}University of California, Berkeley \\
\textsuperscript{3}The Hong Kong University of Science and Technology \ \ \ \ \
\textsuperscript{4}Carnegie Mellon University \\
\href{mailto:shangbin@cs.washington.edu}{\texttt{shangbin@cs.washington.edu}}
}

\begin{document}
\maketitle
\begin{abstract}
Despite efforts to expand the knowledge of large language models (LLMs), knowledge gaps---missing or outdated information in LLMs---might always persist given the evolving nature of knowledge. In this work, we study approaches to identify LLM knowledge gaps and \emph{abstain} from answering questions when knowledge gaps are present. We first adapt existing approaches to model calibration or adaptation through fine-tuning/prompting and analyze their ability to abstain from generating low-confidence outputs. Motivated by their failures in self-reflection and over-reliance on held-out sets,  
we propose two novel approaches that are based on model collaboration, i.e., LLMs probing other LLMs for knowledge gaps, either cooperatively or competitively. 
Extensive experiments with three LLMs on four QA tasks featuring diverse knowledge domains demonstrate that both cooperative and competitive approaches to unveiling LLM knowledge gaps 
achieve up to 19.3\% improvements on abstain accuracy against the strongest baseline. Further analysis reveals that our abstention methods pinpoint failure cases in retrieval augmentation and knowledge gaps in multi-hop reasoning.\footnote{Code and data will be publicly available at \href{https://github.com/BunsenFeng/AbstainQA}{https://github.com/BunsenFeng/AbstainQA}.}
\end{abstract}

\section{Introduction}
LLMs demonstrate impressive capabilities of encoding real-world knowledge in model parameters and leveraging it to aid knowledge-intensive tasks \citep{petroni2019language, brown2020language, yu2023kola}. But when such knowledge is missing or unreliable, they resort to hallucinations \citep{ji2023survey} and biases \cite{feng-etal-2023-pretraining}, while still ``speaking with confidence.'' 
A growing body of work seeks to expand LLM knowledge through retrieval augmentation \citep{guu2020retrieval, borgeaud2022improving, khattab2022demonstrate, shi2023replug, chen2023dense}, search engine integration \citep{nakano2021webgpt, press-etal-2023-measuring}, and multi-LM collaboration \citep{luo2023augmented, Feng2023KnowledgeCF}. However, LLM knowledge gaps might always persist due to the ever-evolving nature of knowledge \citep{kandpal2023large, mallen2023not, de2021editing, hernandez2023measuring, kasai2022realtime}. Consequently, we posit that \emph{abstaining from generating low-confidence outputs should be a part of LLMs' functionality}, and 
ask a crucial research question: \emph{how to identify knowledge gaps in LLMs?} 
Developing and evaluating robust mechanisms to address the abstain problem improves LLM reliability, reduces hallucinations, and mitigates biases due to model uncertainty.


We hypothesize that there are three ways to operationalize the abstain problem: (1) from the \emph{data} perspective, LLMs should abstain when knowledge is missing in training data or curated from unreliable sources;  
(2) from the \emph{modeling} perspective, LLMs should abstain when knowledge is not encoded in model parameters and cannot be inferred; 
(3) from the \emph{usage} perspective, LLMs should abstain when generated outputs would provide an incorrect answer. Since users are directly interacting with and impacted by LLM-generated texts, not model parameters or training data, we formulate and evaluate LLM abstention focusing on the factual correctness of LLM generations. 

Specifically, we focus on \textbf{abstain}ing in \textbf{q}uestion-\textbf{a}nswering (AbstainQA): given an $\mathrm{LLM}$ and a question $\boldsymbol{q}$, we aim to develop an abstain function $f(\mathrm{LLM}, \boldsymbol{q}) \rightarrow \{\textit{true}, \textit{false}\}$ to indicate whether the LLMs should abstain based on the limitations of their internal knowledge \citep{kamath-etal-2020-selective, jiang2021can, whitehead2022reliable}. To facilitate a comprehensive understanding of existing approaches that can be employed for LLM abstention, we first identify 11 methods from previous works and adapt them to incorporate the abstain functionality. We categorize these baselines into (1) \emph{calibration-based}, such as employing temperature scaling \citep{guo2017calibration, jiang2021can} to calibrate models and set a confidence score threshold for abstention; (2) \emph{training-based}, such as employing linear probing on LLMs' hidden layers to evaluate the veracity of generated texts \citep{slobodkin2023curious, azaria2023internal}; (3) \emph{prompting-based}, such as prompting LLMs to request more information before answering questions \citep{Feng2023KnowledgeCF}; and (4) \emph{self-consistency based}, such as generating multiple chains-of-thought \citep{wei2022chain}.

Among the 11 baselines, a majority and often the stronger approaches would require a \emph{held-out set} for training and hyperparameter tuning, potentially harming generalization across knowledge domains. In addition, these approaches often rely on a ``self-reflection'' assumption that \emph{a single LLM} could be employed or adapted to evaluate its own generated texts; however, challenges such as hallucination and confirmation biases cast doubt on this assumption \citep{ji2023survey, xie2023adaptive}.

We propose \textbf{multi-LLM collaboration-based approaches}, enabling robust LLM abstention through multi-LLM collaboration to reflect on generated text in cooperative or competitive settings while removing the need for held-out sets. For multiple LLMs to work in \emph{cooperation} (\textsc{Cooperate}), 
the LLM employs other models to provide feedback on the proposed answer and reasoning, and synthesizes the outputs into an overall abstain decision. For a \emph{competitive} setting (\textsc{Compete}), the LLM 
is challenged by other LLMs with conflicting knowledge, producing an abstain decision based on whether it continues to prefer the proposed answer in the presence of conflicting evidence.

We evaluate the baselines and proposed collaboration approaches for abstention with three LLMs, on four knowledge-intensive QA tasks spanning diverse knowledge domains and reasoning. Extensive experiments demonstrate that \textsc{Cooperate} and \textsc{Compete} are strong abstain mechanisms that outperform all baselines in 9 of the 12 settings across tasks and models, achieving an improvement of up to 19.3\% in abstain accuracy. Further analysis reveals that the proposed collaboration-based approaches could help identify failure cases of retrieval augmentation, pinpoint knowledge gaps in multi-hop reasoning, and more. Our contributions are (1) a critical evaluation and typology of diverse existing methods, repurposed to identify knowledge gaps in LLMs, as well as (2) two novel, robust multi-LLM collaboration methods to detect LLM knowledge gaps, \textsc{Cooperate} and \textsc{Compete}.

\section{Identifying Knowledge Gaps in LLMs}


We propose to repurpose existing approaches to enable abstain functionality in the face of knowledge gaps in LLMs. We conceptually categorize these approaches into \emph{calibration} (\Sref{subsec:calibration}), \emph{training} (\Sref{subsec:training}),  \emph{prompting} (\Sref{subsec:prompting}), and 
\emph{self-consistency} (\Sref{subsec:self-consistency}) methods. 
Empirically observing that these approaches are dependent on external knowledge/data sets and often suffer from instability, we propose a new category based on \emph{multi-LLM collaboration} (\Sref{subsec:multi-LLM}).  

\subsection{Calibration-Based}
\label{subsec:calibration}
\paragraph{Token Probability}  LLMs are often inherently calibrated to different extents \citep{radford2019language, liang2022holistic} and token probabilities might be employed off-the-shelf. Let $(\boldsymbol{q}, \boldsymbol{\Bar{a}}) \in \mathcal{H}$ be a held-out set of questions and gold answers for hyperparameter tuning, while LLM produces an answer $\boldsymbol{a} = \mathrm{LLM}(\boldsymbol{q})$ with a confidence score based on token probabilities $p(\boldsymbol{a}) \in [0,1]$. We set a threshold $p^*$ by minimizing abstain errors on $\mathcal{H}$, formally:

\begin{figure*}[t]
    \centering
    \includegraphics[width=1\linewidth]{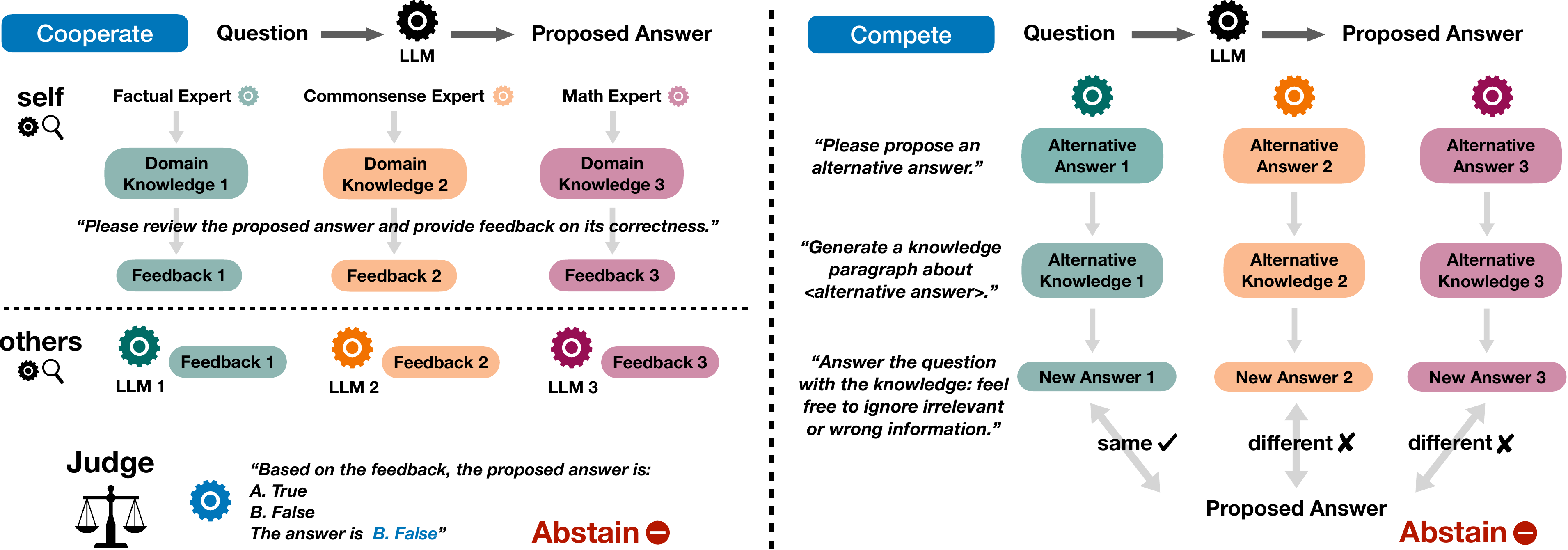}
    \caption{Overview of collaboration-based approaches for LLM abstention: \textsc{Cooperate} and \textsc{Compete}.}
    \label{fig:overview}
\end{figure*}

\begin{align*}
p^* = \argmin_{p' \in [0,1]} \sum_{\boldsymbol{a} = \Bar{\boldsymbol{a}}} \Scale[0.9]{\mathds{1}(p(\boldsymbol{a}) < p')} + \sum_{\boldsymbol{a} \neq \Bar{\boldsymbol{a}}} \Scale[0.9]{\mathds{1}(p(\boldsymbol{a}) \geq p')}
\end{align*}
where $\mathds{1}$ denotes the indicator function. At inference time, the LLM abstains when $p(\boldsymbol{a}) < p^*$.

\paragraph{Temperature Scaling} Before setting a threshold $p^*$ on confidence scores, temperature scaling \citep{guo2017calibration, jiang2021can} first reshapes the token probability distributions by introducing a scalar temperature $\tau > 0$. Given an answer probability distribution $\mathbf{a}$, we transform it into $\mathrm{softmax}(\mathbf{a} / \tau)$, then optimize $\tau$ on the held-out set $\mathcal{H}$ with cross-entropy loss to obtain $\tau^*$. A probability threshold $p^*$ is then set over $\{\mathrm{softmax}(\mathrm{LLM}(\boldsymbol{q}) / \tau^*)\}_{\boldsymbol{q} \in \mathcal{H}}$ and applied at inference time.

\paragraph{Ask for Calibration} In addition to token probabilities, LLMs could also verbalize confidence scores: we follow \citet{tian2023just} to obtain confidence scores $p(\boldsymbol{a})$ by employing a two-step process: 1) generating an answer with the LLM and 2) eliciting a confidence score by prompting with \textit{``Provide the probability that your guess is correct.''}\footnote{Full prompt templates are in Appendix \ref{sec:experiment_details}.} A threshold $p^*$ is similarly set over verbalized scores and applied at inference time.

\subsection{Training-Based}
\label{subsec:training}

\paragraph{Hidden Layers} Previous works demonstrate that probing LLMs' hidden layer representations might yield insights into their knowledge and factuality \citep{slobodkin2023curious, azaria2023internal, ch2023androids}. We extract the hidden embeddings $\mathbf{e}_{\boldsymbol{q}} = \mathrm{LLM}\textit{-hidden}(\boldsymbol{q})$, then conduct linear probing to predict $\ell \in \{\textit{accept}, \textit{reject}\}$ through $p(\ell \mid \boldsymbol{q}) \propto \exp(\mathrm{FFN}(\mathbf{e}_{\boldsymbol{q}}))$. We optimize the cross-entropy loss on the held-out set $\mathcal{H}$. At inference time, the LLM should abstain if $\argmax_{\ell} p(\ell \mid \boldsymbol{q})$ indicates that the answer should be rejected.

\paragraph{External Verifier} Training an external verifier to judge the veracity of LLM-generated text has shown great potential in math reasoning \citep{cobbe2021training} and general knowledge domains \citep{kadavath2022language, cohen2023crawling}. We train an encoder-based language model $\mathrm{LM}\textit{-enc}$ and employ the $\mathrm{[CLS]}$ token for classification. Formally, $p(\ell \mid \boldsymbol{q}) \propto \exp(\mathrm{LM}\textit{-enc}(\boldsymbol{q};\mathrm{LLM}(\boldsymbol{q})))$ is optimized on $\mathcal{H}$ with the cross-entropy loss and an abstain decision is achieved through $\argmax_{\ell} p(\ell \mid \boldsymbol{q})$ at inference time. Here both the question and the generated answer are employed to train the external verifier, different from the \emph{hidden layers} approach where only the intermediate encodings of the question are used for linear probing.

\paragraph{Instruction Tuning} A promising way to adapt LLMs is through tuning LLMs on instruction triplets in the format of $\{\textit{instruction} | \textit{input} | \textit{output}\}$ \citep{ouyang2022training}. Here we investigate whether abstaining could be baked into existing LLMs through instruction tuning. We first employ an instruction to encourage LLMs to abstain when necessary.
Given the held-out set $\mathcal{H}$, we first evaluate the correctness of the generated answer $\boldsymbol{a} = \mathrm{LLM}(\boldsymbol{q})$: if it is correct, we use $\{\textit{instruction}|\boldsymbol{q}|\boldsymbol{a}\}$ as an instruction tuning instance; otherwise, we replace the incorrect answer $\boldsymbol{a}$ with an abstain message and train on $\{\textit{instruction}|\boldsymbol{q}|\textit{abstain message}\}$. In this way, the LLM abstains if the abstain message is generated at inference time for a given $\boldsymbol{q}$.

\subsection{Prompting-Based}
\label{subsec:prompting}

\paragraph{Self-Reflect} Previous studies show that LLMs might have preliminary capabilities of judging and evaluating their own answer \citep{kadavath2022language}. To this end, we prompt the LLM to self-reflect directly after its generated answer with \textit{``The above answer is: A. True B. False''}. LLMs should abstain when they deem the generated answer as false.

\paragraph{More Information} Since the internal knowledge of LLMs is often incomplete and noisy, existing works explore how to selectively request external information to aid LLM generation \citep{xu2023recomp, asai2023self, Feng2023KnowledgeCF}. Following \citet{Feng2023KnowledgeCF}, we append a prompt about whether more information is needed to answer a given question: \textit{``Do you need more information to answer this question? (Yes or No)''}. The LLM should abstain if more information is requested.

\paragraph{Generate and Match} There might be gaps between LLM generation and ``understanding'' \citep{west2023generative}, which leads to hallucinations when questions are presented in a multiple-choice format \citep{fu2023generate}. Instead of directly choosing from one of the options, we instruct the LLM to first generate an answer \emph{without} access to potential options. We then prompt the LLM to evaluate whether the generated answer matches any of the options: the LLM should abstain if there's no match between the generated answer and options.

\subsection{Consistency-Based}
\label{subsec:self-consistency}

\paragraph{None-of-the-Above} Previous works have investigated LLMs' robustness to possibilities of \emph{non-of-the-above} (\emph{nota}) in QA tasks \citep{kadavath2022language, ding2023knowledge}. We propose an abstain mechanism by adding an extra \emph{nota} option to every question: the LLM should abstain if \emph{nota} is selected as the answer, indicating its low consistency in answering the question.

\paragraph{Self-Consistency Threshold} Self-consistency \citep{wang2022self} aims to generate multiple chain-of-thought \citep{wei2022chain} reasoning paths and conduct majority voting to determine the answer. We develop an abstain mechanism by proposing \emph{plurality index}: given a question $\boldsymbol{q}$ along with the $k$ generated reasoning paths and answers $\{(\boldsymbol{a}_i, \boldsymbol{p}_i)\}_{i = 1,\ldots,k}$, the plurality index is defined as:

\begin{align*}
    \mathrm{plu}(\mathrm{LLM}, \boldsymbol{q}, k) = \max_{\boldsymbol{a}_i} \sum_{j = 1,\ldots,k} \mathds{1}(\boldsymbol{a}_j = \boldsymbol{a}_i)
\end{align*}
where it indicates how large the plurality is among all $k$ reasoning paths. A threshold $\tau^* \in [0,1]$ is then optimized on $\mathcal{H}$ and applied at inference time to abstain when $\mathrm{plu}(\mathrm{LLM}, \boldsymbol{q}, k) < \tau^* \cdot k$.

\begin{table*}[t]
    \centering
    \setlength{\tabcolsep}{1pt}
    \renewcommand{\arraystretch}{0.8}
    \resizebox{0.9\textwidth}{!}{
    \begin{tabular}{lcccc|cccc|cccc|cccc|}
         \toprule[1.5pt]
         \multirow{2}{*}{\textbf{Method}} & \multicolumn{4}{c}{\textbf{MMLU}} & \multicolumn{4}{c}{\textbf{K-Crosswords}} & \multicolumn{4}{c}{\textbf{Hellaswag}} & \multicolumn{4}{c}{\textbf{Propaganda}} \\ 
          & R-Acc & ER & A-Acc & A-F1 & R-Acc & ER & A-Acc & A-F1 & R-Acc & ER & A-Acc & A-F1 & R-Acc & ER & A-Acc & A-F1 \\ \midrule[0.75pt]
         \multicolumn{17}{c}{\textit{\ \ \textbf{\textsc{Mistral-7B}}} }\\ \midrule[0.75pt]
         \textcolor{NavyBlue}{\textsc{Probs}} & .570 & .109 & .608 & .456 & .251 & -.351 & .397 & .422 & .456 & -.041 & .599 & .659 & .337 & -.150 & .590 & .680 \\
         \textcolor{NavyBlue}{\textsc{Temp.}} & .565 & .104 & .601 & .426 & .250 & -.452 & .303 & .182 & .451 & -.047 & .601 & .659 & \underline{.340} & -.170 & .585 & .661 \\
         \textcolor{NavyBlue}{\textsc{Ask Cali.}} & .648 & .141 & .639 & .647 & .233 & \underline{-.023} & .705 & .825 & .455 & -.017 & .616 & .734 & .231 & \underline{-.035} & \underline{.705} & \underline{.824} \\
         \textcolor{DarkOrchid}{\textsc{Hidden}} & .449 & -.085 & .420 & .137 & .104 & -.788 & .107 & .009 & .369 & -.198 & .424 & .336 & .223 & -.510 & .240 & .084 \\
         \textcolor{DarkOrchid}{\textsc{Verifier}} & .568 & .083 & .586 & .534 & .208 & -.083 & \bf .805 & \bf .889 & .405 & -.080 & .550 & .628 & \bf .448 & \bf -.015 & \bf .760 & \bf .853 \\
         \textcolor{DarkOrchid}{\textsc{Instruct}} & \underline{.709} & \underline{.198} & \underline{.688} & \underline{.693} & - & - & - & - & \underline{.616} & \underline{.075} & \underline{.701} & \underline{.771} & - & - & - & - \\
         \textcolor{Dandelion}{\textsc{Reflect}} & .498 & -.002 & .495 & .477 & .111 & -.686 & .203 & .207 & .371 & -.150 & .477 & .500 & .216 & -.275 & .490 & .602 \\
         \textcolor{Dandelion}{\textsc{MoreInfo}} & .504 & .006 & .500 & .216 & \underline{.272} & -.455 & .273 & .003 & .379 & -.227 & .396 & .120 & .219 & -.540 & .240 & .073 \\
         \textcolor{Dandelion}{\textsc{Gen+Match}} & .511 & .021 & .515 & .129 & .111 & -.694 & .193 & .188 & .377 & -.174 & .458 & .415 & .226 & -.545 & .230 & .013 \\
         \textcolor{OliveGreen}{\textsc{NOTA}} & .516 & .029 & .528 & .163 & .098 & -.797 & .102 & .011 & .371 & -.244 & .387 & .105 & .259 & -.410 & .340 & .267 \\
         \textcolor{OliveGreen}{\textsc{SC Thres.}} & .604 & .140 & .641 & .551 & .210 & -.090 & \underline{.793} & \underline{.880} & .493 & -.004 & .614 & .713 & .273 & -.100 & .685 & .799 \\ \midrule[0.75pt]
         \textcolor{Maroon}{\textsc{Coop-self}} & .571 & .059 & .564 & .601 & .260 & -.437 & .313 & .311 & .406 & -.042 & .601 & .719 & .297 & -.095 & .680 & .797 \\
         \textcolor{Maroon}{\textsc{Coop-others}} & .688 & \bf .213 & \bf .712 & .692 & .266 & \bf -.022 & .761 & .852 & \bf .626 & \bf .092 & \bf .725 & \bf .783 & .182 & -.140 & .625 & .757 \\
         \textcolor{Maroon}{\textsc{Compete}} & \bf .735 & .140 & .640 & \bf .700 & \bf .289 & -.129 & .597 & .722 & .573 & .032 & .658 & .766 & .302 & -.055 & .700 & .805 \\ \midrule[0.75pt]
         \multicolumn{17}{c}{\textit{\ \ \textbf{\textsc{LLaMA2-70B}}} }\\ \midrule[0.75pt]
         \textcolor{NavyBlue}{\textsc{Ask Cali.}} & .624 & .025 & .435 & .568 & - & - & - & - & \bf .944 & .032 & .498 & .649 & - & - & - & - \\
         \textcolor{DarkOrchid}{\textsc{Hidden}} & .473 & -.018 & .400 & .446 & .282 & -.265 & .423 & .466 & .507 & .007 & .497 & .479 & - & - & - & - \\
         \textcolor{DarkOrchid}{\textsc{Verifier}} & .665 & .201 & .609 & .511 & \underline{.443} & -.056 & .634 & .694 & .522 & .009 & .504 & .617 & .259 & \underline{-.065} & \underline{.755} & \bf .855 \\
         \textcolor{DarkOrchid}{\textsc{Instruct}} & \underline{.745} & .216 & .628 & \bf .640 & .288 & \bf .024 & .606 & \bf .772 & .475 & -.008 & .487 & .615 & - & - & - & - \\
         \textcolor{Dandelion}{\textsc{Reflect}} & .616 & .121 & .529 & .469 & .383 & -.155 & .528 & .537 & .569 & .076 & .560 & .530 & .198 & -.520 & .275 & .225 \\
         \textcolor{Dandelion}{\textsc{MoreInfo}} & .590 & .177 & .590 & .042 & .306 & -.382 & .308 & .020 & .516 & .032 & .518 & .044 & - & - & - & - \\
         \textcolor{Dandelion}{\textsc{Gen+Match}} & .667 & .050 & .450 & .560 & .248 & -.111 & .573 & .708 & .484 & -.004 & .477 & .614 & .082 & -.205 & .620 & .759 \\
         \textcolor{OliveGreen}{\textsc{NOTA}} & .592 & .167 & .583 & .181 & .323 & -.295 & .388 & .280 & .516 & .027 & .522 & .236 & .185 & -.580 & .225 & .124 \\
         \textcolor{OliveGreen}{\textsc{SC Thres.}} & .684 & \underline{.247} & \underline{.656} & .534 & .426 & -.090 & .590 & .617 & .667 & .100 & .590 & \underline{.655} & \bf .412 & \bf -.030 & \bf .760 & \underline{.852} \\ \midrule[0.75pt]
         \textcolor{Maroon}{\textsc{Coop-self}} & .615 & .150 & .550 & .400 & \bf .463 & \underline{-.030} & \underline{.640} & .714 & .649 & \underline{.110} & \underline{.600} & .643 & .222 & -.250 & .500 & .615 \\
         \textcolor{Maroon}{\textsc{Coop-others}} & .694 & \bf .262 & \bf .676 & .562 & .402 & -.063 & .636 & .757 & \underline{.700} & \bf .238 & \bf .704 & \bf .677 & \underline{.329} & -.125 & .675 & .774 \\
         \textcolor{Maroon}{\textsc{Compete}} & \bf .782 & .148 & .552 & \underline{.608} & .323 & -.080 & \bf .642 & \underline{.760} & .611 & .047 & .525 & .625 & .161 & -.210 & .595 & .729 \\ 
         \midrule[0.75pt]
         \multicolumn{17}{c}{\textit{\ \ \textbf{\textsc{ChatGPT}}} }\\ \midrule[0.75pt]
         \textcolor{NavyBlue}{\textsc{Probs}} & .774 & .421 & .715 & .457 & .600 & .187 & .587 & .122 & .750 & .278 & .599 & .476 & .333 & -.015 & .625 & .765 \\
         \textcolor{NavyBlue}{\textsc{Temp.}} & .769 & .419 & .716 & .452 & .616 & .214 & .619 & .216 & .750 & .278 & .595 & .468 & .250 & -.010 & .630 & .772 \\
         \textcolor{NavyBlue}{\textsc{Ask Cali.}} & .694 & .385 & .690 & .006 & .601 & .202 & .601 & .010 & .672 & .344 & .672 & .006 & .444 & -.015 & .580 & .712 \\
         \textcolor{DarkOrchid}{\textsc{Verifier}} & .788 & .301 & .599 & .483 & - & - & - & - & .667 & .305 & .634 & .120 & .441 & -.035 & .570 & .672 \\
         \textcolor{DarkOrchid}{\textsc{Instruct}} & .840 & .422 & \bf .748 & \underline{.579} & .752 & \bf .310 & \bf .709 & .627 & .817 & \bf .429 & \bf .751 & \underline{.614} & .605 & .045 & .645 & .744 \\
         \textcolor{Dandelion}{\textsc{Reflect}} & .752 & .336 & .630 & .411 & .784 & \underline{.239} & .641 & \underline{.633} & .754 & \underline{.377} & \underline{.701} & .487 & .571 & .015 & .615 & .742 \\
         \textcolor{Dandelion}{\textsc{MoreInfo}} & .721 & .246 & .546 & .390 & .605 & .145 & .553 & .380 & .675 & .224 & .548 & .339 & .416 & -.145 & .470 & .293 \\
         \textcolor{Dandelion}{\textsc{Gen+Match}} & .737 & .350 & .652 & .383 & .660 & .083 & .486 & .550 & .712 & .182 & .506 & .447 & .365 & -.115 & .490 & .568 \\
         \textcolor{OliveGreen}{\textsc{NOTA}} & .719 & .389 & .692 & .260 & .644 & .163 & .565 & .480 & .689 & .307 & .628 & .268 & .400 & -.120 & .485 & .488 \\
         \textcolor{OliveGreen}{\textsc{SC Thres.}} & .766 & \underline{.424} & \underline{.743} & .447 & .637 & .216 & .622 & .382 & .749 & .366 & .688 & .468 & .399 & -.160 & .440 & .309 \\ \midrule[0.75pt]
         \textcolor{Maroon}{\textsc{Coop-self}} & \underline{.841} & \bf .436 & .726 & .578 & \underline{.794} & .175 & \underline{.646} & \bf .646 & \underline{.878} & .344 & .670 & \bf .628 & \underline{.684} & \underline{.070} & \bf .710 & \bf .802 \\
         \textcolor{Maroon}{\textsc{Coop-others}} & .780 & .362 & .660 & .479 & .659 & .109 & .509 & .536 & .790 & .350 & .676 & .565 & \bf .790 & \bf .321 & .647 & .543 \\
         \textcolor{Maroon}{\textsc{Compete}} & \bf .947 & .306 & .602 & \bf .583 & \bf .875 & .034 & .441 & .589 & \bf .939 & .172 & .490 & .545 & .611 & .040 & \underline{.670} & \underline{.795} \\
         \bottomrule[1.5pt]
    \end{tabular}
    }
    \caption{Performance of abstain strategies on four datasets and three LLMs. Best results in \textbf{bold} and second best in \underline{underline}. Approaches are color-coded per category: \textcolor{NavyBlue}{calibration}, \textcolor{DarkOrchid}{training}, \textcolor{Dandelion}{prompting}, \textcolor{OliveGreen}{consistency}, and \textcolor{Maroon}{collaboration}. Certain incompatible cases, \emph{e.g.} \textsc{embedding} with the black-box \textsc{ChatGPT}, are omitted. ``-'' indicates that this approach fails to produce meaningful abstain decisions: almost always abstain, didn't follow instructions, etc. \textsc{Cooperate} and \textsc{Compete} achieve the best performance in 9 of the 12 settings in terms of reliable accuracy.}
    \label{tab:big}
\end{table*}

\subsection{Proposed: Multi-LLM Collaboration}
\label{subsec:multi-LLM}

Existing approaches often adapt an LLM to evaluate/reflect on its own generated texts, while challenges such as confirmation bases and hallucination would result in unreliable self-evaluation \citep{kadavath2022language, ji2023survey, xie2023adaptive}. In addition, the stronger approaches rely on held-out sets and result in weaker generalization across knowledge domains. We propose to enable robust LLM abstention through \textbf{multi-LLM collaboration}, i.e., employing multiple LLMs that interact with each other to evaluate the veracity of generated texts and make abstain decisions (Figure \ref{fig:overview}). 

\paragraph{Cooperate} LLMs often come with varying knowledge coverage with the potential of complementing each other \citep{yu2023kola, du2023improving, Bansal2024LLMAL}, while the internal knowledge of one LLM could also be diversified and specialized \citep{kang2023self, si2023getting}. We tap into this knowledge variation in the \textsc{Cooperate} approach by having different expert LLMs 
generate feedback on LLM-proposed answers and employing an LLM as the final judge to decide whether to abstain. Formally, given question $\boldsymbol{q}$ and answer $\boldsymbol{a} = \mathrm{LLM}(\boldsymbol{q})$, we obtain a set of natural language feedback $\{\boldsymbol{f}_1,\ldots,\boldsymbol{f}_k\}$ from expert LLMs and employ a judge LLM to summarize and synthesize the feedbacks into an abstain decision $\mathrm{LLM}(\boldsymbol{q}, \boldsymbol{a}, \{\boldsymbol{f}_1,\ldots,\boldsymbol{f}_k\}) \rightarrow \{\textit{accept}, \textit{reject}\}$.

We employ two modes to generate feedback: for \emph{self}, the same LLM is specialized into experts on domains $\{d_1, \ldots, d_k\}$ through prompting-based self-specialization: we prompt the LLM to generate a knowledge passage $\textit{knowledge}_i$ about $\boldsymbol{q}$ with a focus on domain $d_i$. A domain-specific feedback is then generated by prepending the knowledge passage: $\boldsymbol{f}_i = \mathrm{LLM}(\textit{knowledge}_i, \boldsymbol{q}, \boldsymbol{a})$, and prompting the model to respond as a reviewer. For \emph{others}, we use separate LLMs $\{\mathrm{LLM}_1, \ldots, \mathrm{LLM}_k\}$ to provide feedback $\boldsymbol{f}_i = \mathrm{LLM}_i(\boldsymbol{q}, \boldsymbol{a})$, aiming to identify the knowledge gaps in each other and complement the abstain process.

\paragraph{Compete} LLMs have varying preferences for knowledge conflicts, \emph{i.e.} when there is a conflict between LLM's internal parametric knowledge and external information provided in the prompt \citep{xie2023adaptive, wang2023resolving}. We hypothesize that an LLM should abstain when it is impacted by conflicting information generated by other LLMs and does not stick to the original answer. Concretely, for question $\boldsymbol{q}$ and LLM-generated answer $\boldsymbol{a} = \mathrm{LLM}(\boldsymbol{q})$, we prompt other LLMs to generate $k$ alternative answers $\{\boldsymbol{a}^{'}_{1}, \ldots, \boldsymbol{a}^{'}_{k}\}$ along with a knowledge passage $\textit{knowledge}_i$ about $\boldsymbol{a}^{'}_{i}$. We then instruct the LLM to answer $\boldsymbol{q}$ again with conflicting information prepended: $\Tilde{\boldsymbol{a}}_i = \mathrm{LLM}(\textit{knowledge}_i,\boldsymbol{q})$. If $\boldsymbol{a} = \Tilde{\boldsymbol{a}}_i$, the LLM sticks to the original answer and should be thus accepted; otherwise, the LLM is swayed by conflicting information generated by other LLMs, betraying its low confidence. This process is repeated for $k$ alternative answers and the LLM should abstain if the answer changes in a majority of cases.
\section{Experiment Settings}
\paragraph{Models} We evaluate LLM abstention baselines and our proposed collaboration-based approaches with three LLMs featuring varying sizes and openness: \emph{Mistral-7B} \citep{jiang2023mistral}, \emph{LLaMA2-70B} \citep{touvron2023llama}, and \emph{ChatGPT}. We posit that a robust abstain mechanism should ideally work for all LLMs, weak and strong. We set the default sampling temperature to 0.1, and employ 0.7 where multiple runs are required.

\paragraph{Tasks and Datasets} We evaluate LLM abstention with four datasets spanning diverse knowledge domains and reasoning scenarios: 1) \emph{MMLU} \citep{hendrycks2020measuring} is a multiple-choice dataset for general knowledge QA; 2) \emph{Knowledge Crosswords} \citep{ding2023knowledge} is a structured QA dataset that focuses on multi-hop and compositional knowledge reasoning; 3) \emph{Hellaswag} \citep{zellers2019hellaswag} is a natural language inference dataset that tests commonsense knowledge and reasoning; 4) \emph{Propaganda} \citep{piskorski2023semeval} tasks LLMs with identifying the 23 propaganda tactics in a long news article based on their internal knowledge. The datasets are all in English. We create splits of held-out sets and test sets for the four datasets with details in Appendix \ref{subsec:dataset_details}. We employ LLMs to answer these questions with zero-shot prompting and greedy decoding to obtain the ground truth of whether the LLM ``knows'' the answer and whether it should abstain. We posit that a robust abstain mechanism should work for knowledge tasks across domains and reasoning contexts.
\vspace*{-25pt}

\begin{wrapfigure}{l}{0.4\linewidth}
    \vspace*{-15pt}
    \centering
    \includegraphics[width=1\linewidth]{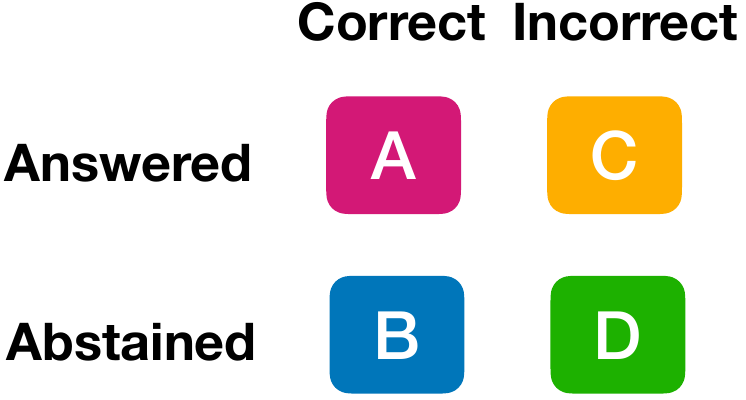}
    \caption{Four outcomes of AbstainQA.}
    \label{fig:ABCD}
    \vspace*{-15pt}
\end{wrapfigure}

\paragraph{Evaluation Metrics} We illustrate the four scenarios under AbstainQA in Figure \ref{fig:ABCD} and employ the following evaluation metrics. 
(1) \emph{Reliable Accuracy} (R-Acc): $\frac{\mathcal{A}}{\mathcal{A}+\mathcal{C}}$ indicates to what extent could LLM-generated answers (not abstained) be trusted.
i.e., \emph{out of all questions answered, how many are correct?}  
(2) \emph{Effective Reliability} (ER) \citep{whitehead2022reliable, si2023getting}: $\frac{\mathcal{A}-\mathcal{C}}{\mathcal{A}+\mathcal{B}+\mathcal{C}+\mathcal{D}}$ strikes a balance between reliability and coverage.
i.e., \emph{out of all questions, what proportion more are answered correctly than incorrectly?}
(3) \emph{Abstain Accuracy} (A-Acc): $\frac{\mathcal{A}+\mathcal{D}}{\mathcal{A}+\mathcal{B}+\mathcal{C}+\mathcal{D}}$ evaluates whether the abstain decisions are correct: LLMs should abstain when it would provide an incorrect answer and vice versa. 
(4) \emph{Abstain F1} (A-F1): $\textit{harmonic-mean}(\mathrm{precision},\mathrm{recall})$, where $\mathrm{precision} = \frac{\mathcal{D}}{\mathcal{B}+\mathcal{D}}$ and $\mathrm{recall} = \frac{\mathcal{D}}{\mathcal{C}+\mathcal{D}}$, a balanced metric between reliability and answer coverage.

\section{Results}
\label{sec:results}
We present the performance of abstain mechanisms with three LLMs on four tasks in Table \ref{tab:big}.

\paragraph{\textsc{Cooperate} and \textsc{Compete} are state-of-the-art abstain mechanisms.} Our proposed collaboration-based approaches outperform the strongest baseline in 9 out of the 12 settings (3 models and 4 datasets), achieving an average improvement of 10.1\% on reliable accuracy scores. We find that \textsc{Cooperate} works better with stronger models such as ChatGPT: we hypothesize that the complexities of the multi-step feedback generation warrant a stronger base LLM. \textsc{Compete} emphasizes reliability and greatly avoids wrong answers evident in the high reliable accuracy scores: we observe that LLMs stick to the original answers only in the most confident scenarios, making \textsc{Compete} ideal for abstention when reliability is paramount.

\paragraph{Calibration > Training > Consistency > Prompting} On average, the four types of approaches achieved 0.595, 0.576, 0.553, and 0.475 A-Acc scores. While simple prompting-based approaches struggle to elicit self-reflection, our proposed \textsc{Cooperate} and \textsc{Compete} promote collaborative abstention and greatly improve performance. Among all baselines, \emph{instruction tuning} and \emph{self-consistency threshold} stand out as two strong approaches: however, they both rely on a held-out set for training and hyperparameter tuning: we further investigate their generalizability in Section \ref{sec:analysis}.

\paragraph{Abstain capabilities potentially correlate with base LLM utility.} Abstain accuracy (A-Acc) is a metric independent of LLMs' underlying performance on a QA dataset and solely evaluates abstention quality. For A-Acc, \emph{Mistral-7B}, \emph{LLaMA2-70B}, and \emph{ChatGPT} achieve 0.524, 0.537, and 0.616 A-Acc scores on average across approaches and datasets, which aligns well with their relevant superiority on standard benchmarks. In addition, our proposed collaboration-based approaches also work better with the strongest \emph{ChatGPT}, while also improving the abstention performance of weaker LLMs over baselines. 

\paragraph{Going beyond generic QA sheds light on the limitations of existing approaches.} While most baselines could function properly on MMLU, some collapse (``-'' in Table \ref{tab:big}) in other knowledge domains or reasoning contexts: \textsc{Ask Calibrate}, among others, struggle to follow instructions when the prompt context is long in detecting propaganda tactics; \textsc{Instruction Tuning} collapses and produces 100\% abstention due to imbalances in the held-out set in K-Crosswords where problems are much harder and the LLM should mostly abstain; calibration approaches struggle to set a good threshold when the model's base performance is too low. However, our proposed collaboration-based approaches work well with all four tasks, especially with the strongest \emph{ChatGPT} language model, indicating its robustness and broad compatibility for knowledge-intensive tasks across domains and reasoning contexts.

\section{Analysis}
\label{sec:analysis}

\begin{figure}[t]
    \centering
    \includegraphics[width=\linewidth]{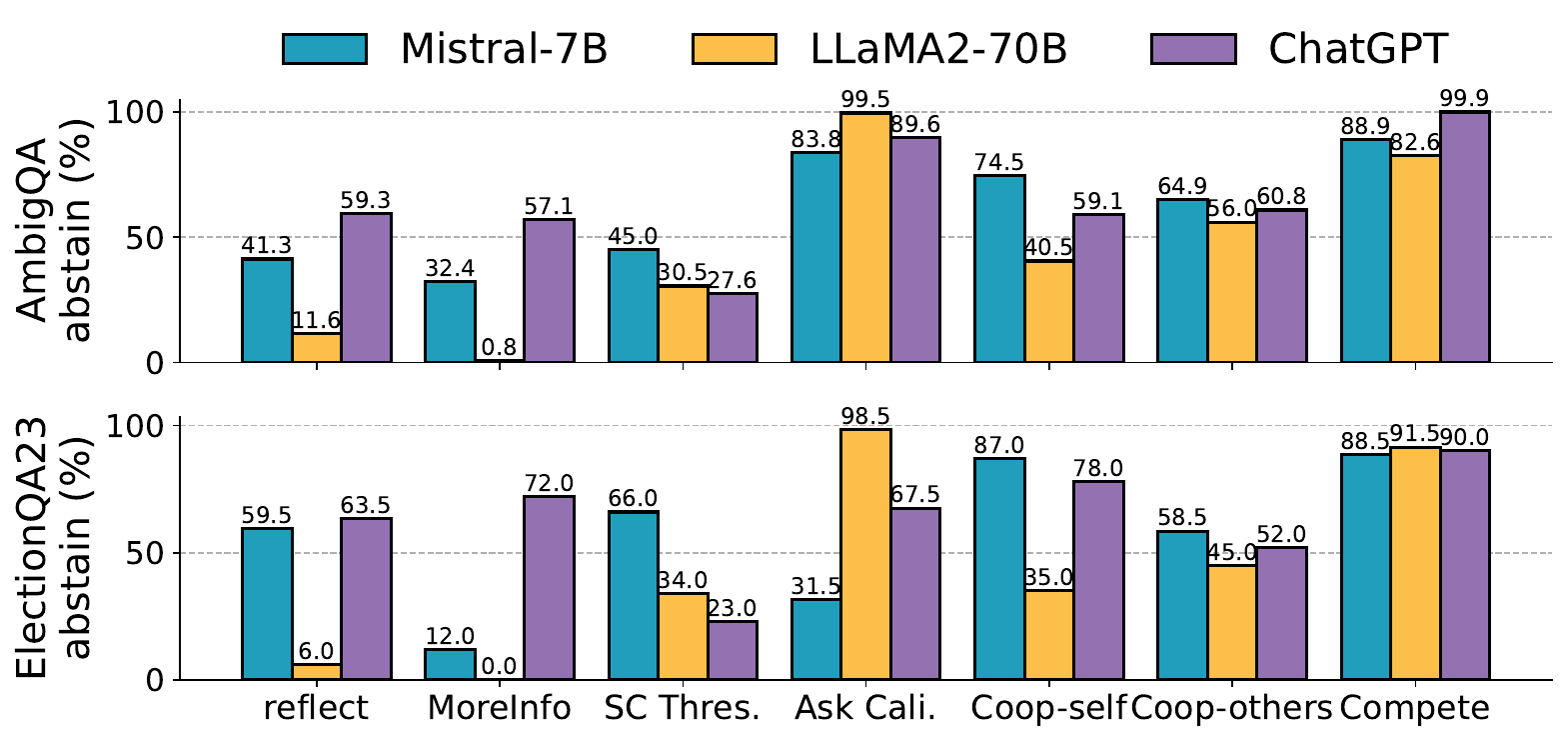}
    \caption{Performance of abstain mechanisms in the \emph{abstain absolute} scenarios where the LLM should abstain for 100\% of questions. \textsc{Compete} achieves the highest abstention rate on average across LLMs and datasets.}
    \label{fig:abstain_absolute}
\end{figure}

\begin{figure*}[t]
    \centering
    \includegraphics[width=1\linewidth]{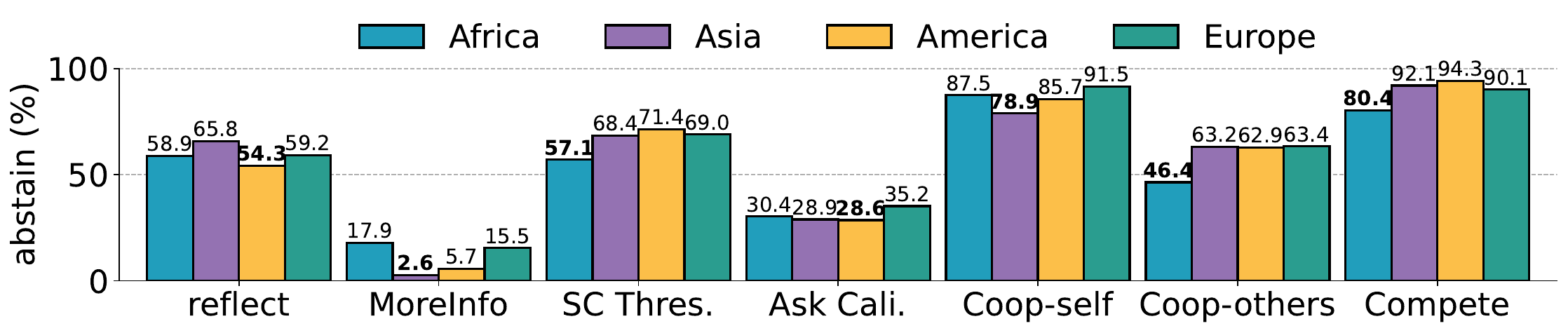}
    \caption{Abstain rate on Election23QA with \emph{Mistral-7B} divided into where the election takes place. The lowest abstain rate for each approach among continents is highlighted in \textbf{bold}.}
    \label{fig:fairness}
\end{figure*}

\paragraph{Abstain Absolute} Aside from QA datasets where abstain decisions should be made based on the potential correctness of answers, we investigate scenarios where LLMs should 100\% abstain. Specifically, we employ AmbigQA \citep{min2020ambigqa}, where LLMs should always abstain from ambiguous and underspecified questions. We also curate ElectionQA23, a QA dataset focusing on elections in 2023 around the globe (details in Appendix \ref{subsec:dataset_details}) and LLMs should abstain due to the temporal mismatch of training data cutoff, \emph{i.e.} if the LLM has only seen training data before 2023. Other potential \emph{abstain absolute} scenarios include known unknowns \citep{amayuelas2023knowledge} and questions with false/underspecified premises \citep{patidar2022have, hu2023won}. Figure \ref{fig:abstain_absolute} demonstrates that our proposed \textsc{Compete} consistently abstains the most in \textit{abstain absolute} scenarios across three LLMs and two datasets: we observe that the proposed alternative answers in \textsc{Compete} often provide contradictory contexts, thus the LLMs successfully abstain due to the knowledge conflicts. We further divide the ElectionQA23 dataset into subsets of which continent is the election held in and present model performance in Figure \ref{fig:fairness}. It is illustrated that LLMs tend to abstain less and assume the outcomes of future elections for Africa and Asia, indicating fairness concerns of abstain mechanisms that might underserve marginalized communities and underdeveloped countries.


\begin{figure}[t]
    \centering
    \includegraphics[width=\linewidth]{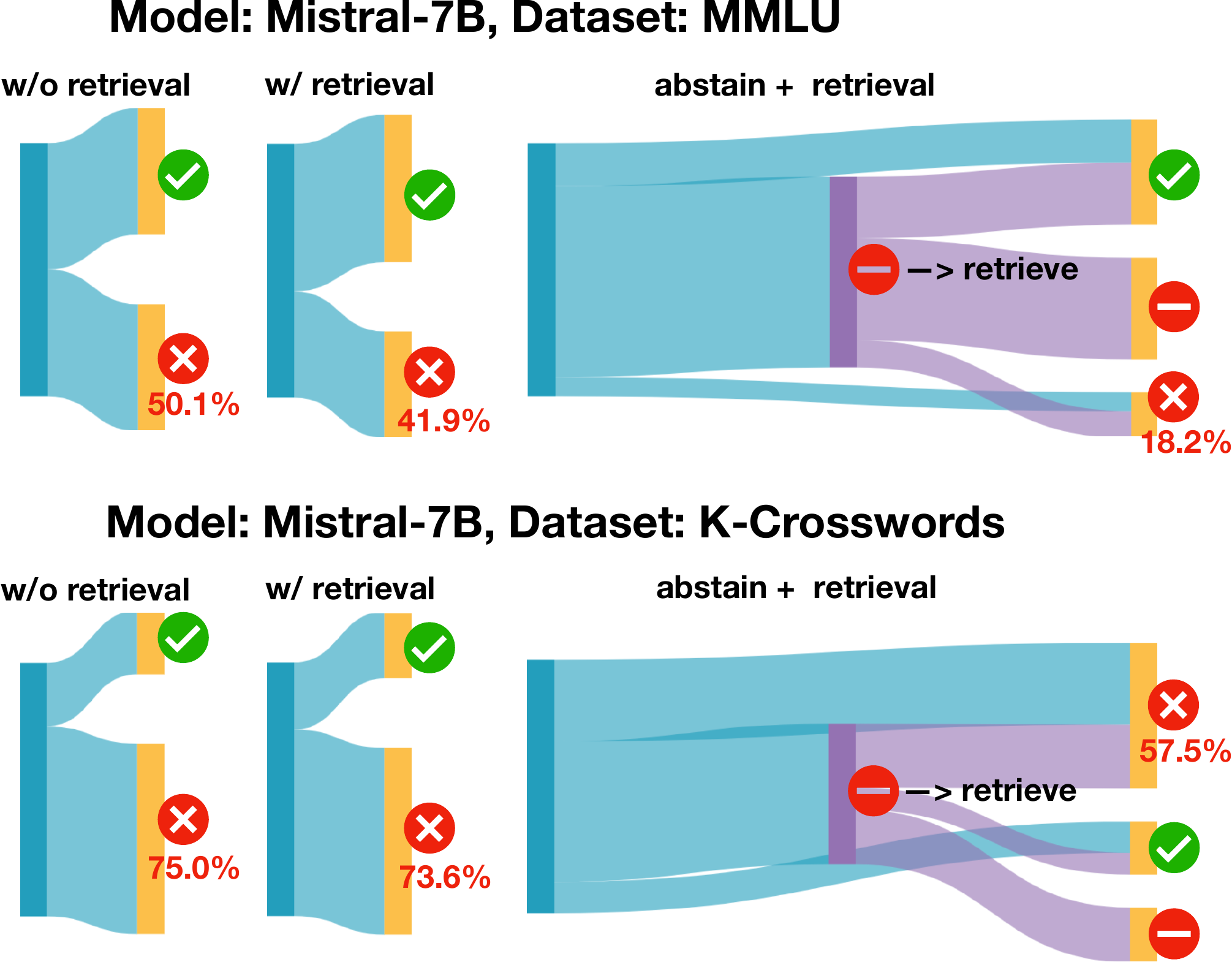}
    \caption{Performance of \textsc{Compete} with \textsc{Mistral-7B} in the two-step abstain and retrieval setting. The proposed abstain-retrieve-abstain pipeline successfully reduces the incorrect rate by at least 21.2\%.}
    \label{fig:retrieval_failure}
\end{figure}

\paragraph{Abstain and Retrieval} Retrieval-augmented language models \citep{khattab2022demonstrate, shi2023replug, baek2023knowledge, yu2023chain, thakur2023nomiracl} have become the go-to approach for augmenting LLMs' knowledge access, while retrieving from an external corpus could also introduce noisy documents and lead to wrong answers \citep{wang2023learning, asai2023self}. We investigate whether abstain mechanisms could work with retrieval, specifically when questions in the MMLU dataset are presented together with a retrieved document from the WikiSearch API. Table \ref{tab:retrieval_impact} demonstrates that while baselines such as \textsc{ask calibrate} are negatively impacted by retrieval, our proposed \textsc{Cooperate} and \textsc{Compete} are in general robust towards additional prepended context through retrieval.

\begin{table}[t]
    \centering
    \resizebox{1\linewidth}{!}{
    \begin{tabular}{lccccc}
         \toprule[1.5pt]
         \textbf{Method} & \textbf{Ret.} & \textbf{R-Acc} & \textbf{ER} & \textbf{A-Acc} & \textbf{A-F1} \\ \midrule[0.75pt]
         \multirow{2}{*}{\textsc{SC Thres.}} & \xmark & 0.604 & 0.140 & 0.641 & 0.551 \\
          & \cmark & 0.634 & 0.160 & 0.652 & 0.611 \\
         \multirow{2}{*}{\textsc{Ask Cali.}} & \xmark & 0.648 & 0.141 & 0.639 & 0.647 \\
          & \cmark & 0.515 & 0.027 & 0.512 & 0.156 \\
          \multirow{2}{*}{\textsc{Coop-self}} & \xmark & 0.571 & 0.059 & 0.564 & 0.601 \\
          & \cmark & 0.567 & 0.048 & 0.536 & 0.589 \\
          \multirow{2}{*}{\textsc{Compete}} & \xmark & 0.735 & 0.140 & 0.640 & 0.700 \\
          & \cmark & 0.720 & 0.128 & 0.618 & 0.670 \\ \bottomrule[1.5pt]
    \end{tabular}
    }
    \caption{Performance of abstain strategies on MMLU with \emph{Mistral-7B} with or without retrieval (Ret.).}
    \label{tab:retrieval_impact}
\end{table}

Building upon the findings that collaboration-based approaches are robust with retrieved contexts, we propose a 2-step abstain-based framework to selectively activate retrieval and identify retrieval failures. Specifically, abstain decisions are first made based on QA without retrieval. If the LLM abstains, retrieval is then requested and the abstain mechanism is applied to QA with the retrieved context: if the LLM still decides to abstain, it indicates that the retrieval has failed to find relevant/helpful documents. We evaluate this framework on MMLU and Knowledge Crosswords datasets while comparing it with no retrieval and full retrieval settings.
Results in Figure \ref{fig:retrieval_failure} demonstrate that this abstain-retrieve-abstain pipeline successfully cuts back the error rate by up to 56.6\%, highlighting the potential of abstain mechanisms in identifying retrieval failures and enhancing the reliability of retrieval-augmented LLMs.

\begin{figure}[t]
    \centering
    \includegraphics[width=\linewidth]{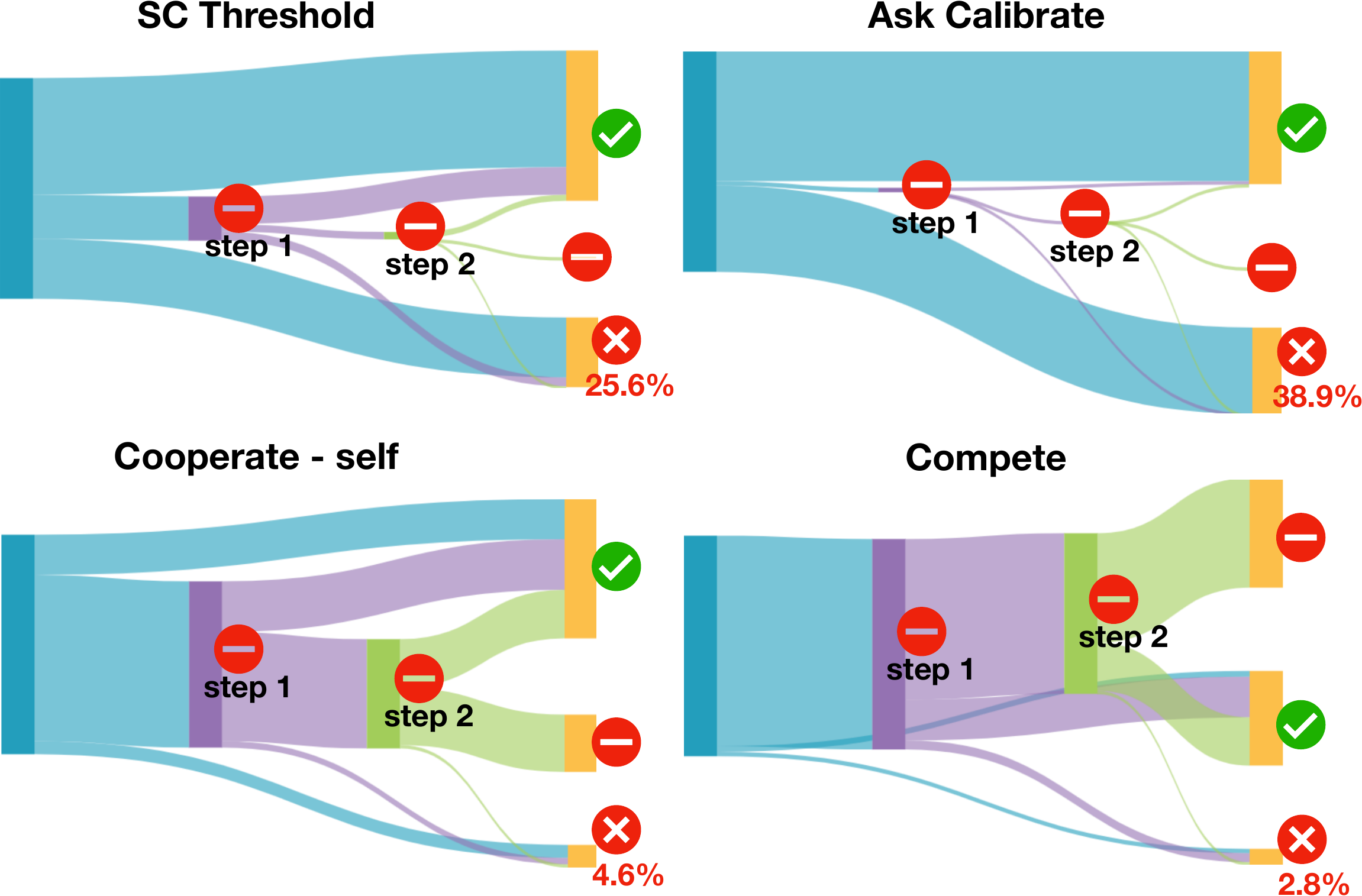}
    \caption{Performance of abstain mechanisms with ChatGPT solving 3-hop questions in K-Crosswords. \textsc{Cooperate} and \textsc{Compete} are better at localizing knowledge gaps in multi-hop reasoning steps and reducing the incorrect rate by at least 82\%.}
    \label{fig:multi-hop}
\end{figure}

\begin{table}[t]
    \centering
    \resizebox{1\linewidth}{!}{
    \begin{tabular}{lcccc}
         \toprule[1.5pt]
         \textbf{Method} & \textbf{MMLU} & \textbf{K-Cross.} & \textbf{Hellaswag} & \textbf{Propaganda} \\ \midrule[0.75pt]
         \textsc{MoreInfo} & 0.4433 & 0.7281 & 0.5830 & 0.6143 \\
         \textsc{Probs} & 0.4253 & 0.6643 & 0.4992 & 0.5302 \\
         \textsc{SC Tres.} & 0.2854 & 0.3428 & 0.2146 & 0.2750 \\
         \textsc{Ask Cali.} & \bf 0.1747 & 0.2664 & \bf 0.2088 & 0.3755 \\
         \textsc{Verifier} & 0.3343 & 0.2000 & 0.3624 & 0.2112 \\
         \textsc{Coop-self} & 0.4740 & \bf 0.1131 & 0.3434 & 0.2263 \\
         \textsc{Coop-others} & 0.3560 & 0.5330 & 0.2164 & 0.2380 \\
         \textsc{Compete} & 0.4484 & 0.3929 & 0.4014 & \bf 0.1935 \\ \bottomrule[1.5pt]
    \end{tabular}
    }
    \caption{Abstain estimated calibration error (ECE) of approaches with \textit{Mistral-7B}, the lower the better. The calibration of abstention scores for approaches could vary significantly across tasks and datasets.}
    \label{tab:ece}
\end{table}

\begin{figure*}[t]
    \centering
    \includegraphics[width=\linewidth]{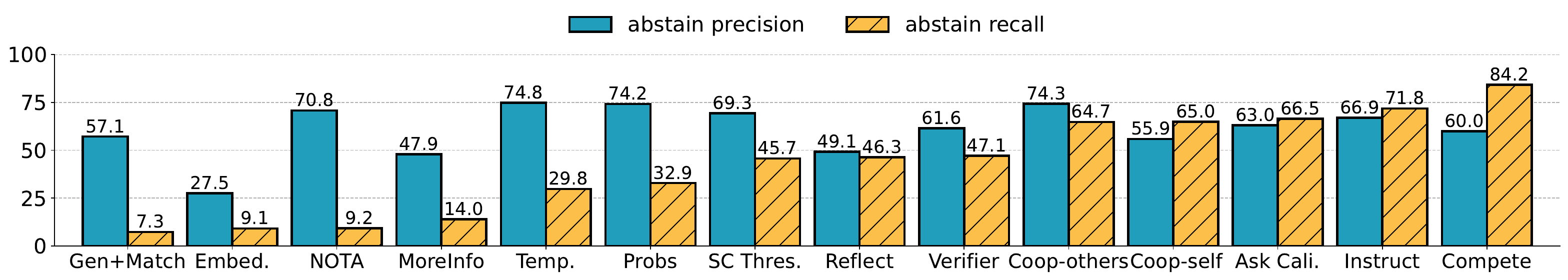}
    \caption{Abstain precision and recall of approaches on MMLU with \emph{Mistral-7B}. While most approaches feature higher precision, \textsc{Compete} achieves the best abstention recall and could be employed where reliability is paramount.}
    \label{fig:precision_and_recall}
\end{figure*}

\paragraph{Abstain in Multi-Hop} In multi-hop knowledge reasoning, LLMs' knowledge is often limited in some, but not all, of the reasoning steps. We investigate whether our proposed abstain strategies could pinpoint the specific steps where LLMs should abstain due to knowledge limitations: we employ the three-hop subset of the K-Crosswords dataset, conduct abstain mechanisms on each of the hops separately, and present results in Figure \ref{fig:multi-hop}. It is demonstrated that our proposed \textsc{Cooperate} and \textsc{Compete} are better in localizing knowledge limitations, with a final error rate of 67.2\% to 81.2\% lower than the strongest baselines.

\paragraph{Abstain ECE} Aside from a binary decision, we posit that abstain mechanisms could also provide a continuous abstain likelihood based on token probabilities, heuristics, and more\footnote{Details in Appendix \ref{sec:experiment_details}.}. We calculate the \emph{abstain ECE}, estimated calibration error \citep{guo2017calibration} but with abstain probabilities and abstain accuracy, on the MMLU dataset and present in Table \ref{tab:ece}. It is demonstrated that the calibration of abstain likelihood generally correlates with the abstain performance, with \textsc{ask calibrate} and our proposed \textsc{Cooperate} and \textsc{Compete} being the top-performing approaches. This indicates that our proposed collaboration-based approaches could also provide well-calibrated abstain likelihood scores where fine-grained decision-making is required.

\paragraph{Precision and Recall} We observe that among the 14 abstain mechanisms, some emphasize answer coverage while others emphasize answer reliability. We illustrate the abstain precision and recall of approaches in Figure \ref{fig:precision_and_recall}: It is illustrated that while most approaches have higher precision than recall and thus lean towards answer coverage, \textsc{Compete} has a much higher abstain recall and stresses answer reliability. We argue that these strategies could be ensembled to achieve balanced abstain decisions or employed based on whether the application scenario should optimize coverage or reliability.

\section{Related Work}
Previous works on knowing what language models know have focused on calibration, prompting, or training. \emph{Calibration-based} approaches attempt to extract confidence scores from models to gauge their uncertainty \citep{sun2022quantifying, kuhn2023semantic, zhou2023batch, liu2023litcab}. Previous works have evaluated \citep{radford2019language, liang2022holistic, tao2023benchmark, he2023investigating} and improved the calibration of language models \citep{desai-durrett-2020-calibration, kong-etal-2020-calibrated, jagannatha-yu-2020-calibrating, kamath-etal-2020-selective, jiang2021can, mielke2022reducing, lin2022teaching, tian2023just, zhou-etal-2023-navigating}, while the calibration quality could vary for different downstream tasks \citep{desai-durrett-2020-calibration, wang-etal-2020-inference, stengel2023calibrated, kalai2023calibrated}. In the setting of AbstainQA where LLMs decide whether to abstain or not given a question, calibration-based approaches would rely on a held-out set to set a threshold over calibrated confidence scores: such a threshold and reliance on a held-out set could jeopardize the generalization of calibration-based approaches across knowledge domains and reasoning contexts.

\emph{Prompting-based} approaches leverage the instruction-following abilities of LLMs and employ instructions to induce self-reflection and gauge whether the generated answer should be trusted. \citet{kadavath2022language} investigates whether a ``none-of-the-above'' option or self-evaluation prompting would induce good estimations of LLMs' internal factuality. \citet{huang2023large} follows existing works \citep{kim2023language, shinn2023reflexion} to evaluate whether LLMs could self-correct their own reasoning with a three-step prompting strategy. \citet{Feng2023KnowledgeCF} prompts LLM to reflect on whether more information is needed for a given question to selectively solicit external knowledge. These approaches, among other prompting-based methods for acknowledging knowledge limitations \citep{wang-etal-2023-chatgpt-defend, si2023getting}, mostly rely on the hope that LLMs could indeed carry out self-reflection through simple prompting, while their ability on this front might be subject to confounders and spurious correlations \citep{kadavath2022language}.

\emph{Training-based} approaches focus on empowering LLMs with the ability to acknowledge uncertainty or abstain through training with special data or objectives. \citet{cobbe2021training} proposes to train an external verifier to evaluate the correctness of LLM-generated answers in math reasoning. A series of works also attempt to probe the hidden representations of LLMs for signals of factuality and hallucination \citep{slobodkin2023curious, azaria2023internal}. Contemporary works have gained interest in improving LLMs' factuality and self-knowledge through instruction tuning and alignment \citep{zhang2023r, yang2023alignment, sun2023aligning, bashlovkina2023trusted}, as well as promoting abstention from a safety perspective \citep{huang2023survey, liu2023trustworthy}. While training-based approaches have shown great potential, they would often require substantial computational resources to fine-tune LLMs and might struggle to generalize across knowledge domains and contexts due to their reliance on the patterns and examples in the training subset.

In this work, we propose \textsc{Cooperate} and \textsc{Compete}, two collaboration-based approaches to enable robust LLM abstention through multi-LLM collaboration.
These approaches have shown great potential in improving LLM abstention in not only direct QA, but also retrieval-augmented QA and multi-hop reasoning scenarios.
\section{Conclusion}
We investigate AbstainQA, a setting where LLMs should abstain from answering questions incorrectly. We curate a taxonomy of 11 abstain baselines across four categories and propose \textsc{Cooperate} and \textsc{Compete}, two novel abstain mechanisms that promote mechanistic reflection through multi-LLM collaboration, in cooperation or competition. Extensive experiments on four datasets demonstrate that \textsc{Cooperate} and \textsc{Compete} advances the state-of-the-art in AbstainQA, with the potential to improve retrieval-augmented LLMs, multi-hop knowledge reasoning, and more.


\section*{Limitations}

While we primarily investigated and evaluated LLM abstention from a knowledge perspective, the problem of abstaining is also important from a safety perspective, preventing LLMs from generating harmful content or biased responses \citep{huang2023survey, liu2023trustworthy}. There are also intersections between the knowledge and safety aspects of abstention, such as abstaining from underspecified questions based on stereotypes of demographic attributes. We plan to expand on this work and propose unified abstain strategies for knowledge and safety scenarios in the future.

The abstain mechanisms and experiments in this work are exhaustive to the extent that our computational budget permits, while we could not cover every single approach and proposal from existing works. Nevertheless, we will make the evaluation data, code, and framework publicly available so that new abstain approaches, as well as datasets, can be seamlessly added to advance LLM abstention research.

Our proposed collaboration-based approaches require prompting multiple LLMs for feedback and could have more computational overhead, while the exact cost would depend on the inference costs of other LLMs. We provide a summary of inference costs in Table \ref{tab:llm_overhead} and Appendix \ref{sec:analysis_cont}. We posit that the value of multi-LLM collaboration for abstention lies in using smaller models to supervise larger models in judging veracity and preventing hallucinations. When we serve a user-facing LLM with hundreds of billions of parameters, our approach enables the incorporation of a few 7B LLMs for abstention guidance and mitigate hallucinations: having a few extra 7B models does not add greatly to the cost of a hundreds-of-billions LLM, while significantly boosting its factuality and reliability. By sampling multiple feedbacks to make abstain decisions, randomness is also introduced and there might be variation across model inference runs.

In this work, we focus on a conceptual ``abstain'' functionality and develop robust approaches to identify the knowledge gaps in large language models. However, ``abstain'' doesn't necessarily have to be part of the user interface of LLMs: it could be an internal component of an LLM-based system, followed by approaches to find better knowledge sources to fill the knowledge gaps or methods to incorporate the uncertainty to model responses to enhance trustworthiness \citep{Zhou2024RelyingOT}. Future work could further explore LLM pipelines where these abstain approaches are one component towards reliable question answering.

We acknowledge the potential impact of test set pollution in the training of the adopted LLMs. While we do not know the exact training composition of ChatGPT and the other two LLMs (Mistral-7B, LLaMA2-70B) are not explicitly trained on test sets, they might still be unintentionally included in the training data. Our results and findings should be thus interpreted with this limitation in mind.

\section*{Ethics Statement}

Figure \ref{fig:fairness} highlights that LLM abstain decisions could also have fairness implications, as they fail to abstain and assume the outcomes of future elections more frequently for African and Asian countries. We similarly expect varying levels of abstain performance on questions and prompts regarding different demographics, communities, and perspectives, potentially due to LLMs' internal social biases \citep{blodgett-etal-2020-language, jin-etal-2021-transferability, bender2021dangers, shaikh-etal-2023-second, feng-etal-2023-pretraining}. We plan to examine the intersections of LLM social biases and their abstention abilities in future work, especially in critical domains such as hate speech and misinformation where failing to abstain from generating problematic content could have grave consequences for marginalized communities.

\section*{Acknowledgements}
This material is based upon work supported by the National Science Foundation under CAREER Grant No.~IIS2142739, NSF Grant No.~IIS2203097, and the DARPA CMO under Contract No. HR001120C0124.  Any opinions, findings and conclusions or recommendations expressed in this material are those of the author(s) and do not necessarily reflect the views of the funding agencies. We also gratefully acknowledge support from Alfred P.~Sloan Foundation Fellowship.

\bibliography{custom}

\begin{thebibliography}{92}
\expandafter\ifx\csname natexlab\endcsname\relax\def\natexlab#1{#1}\fi

\bibitem[{Amayuelas et~al.(2023)Amayuelas, Pan, Chen, and Wang}]{amayuelas2023knowledge}
Alfonso Amayuelas, Liangming Pan, Wenhu Chen, and William Wang. 2023.
\newblock Knowledge of knowledge: Exploring known-unknowns uncertainty with large language models.
\newblock \emph{arXiv preprint arXiv:2305.13712}.

\bibitem[{Asai et~al.(2023)Asai, Wu, Wang, Sil, and Hajishirzi}]{asai2023self}
Akari Asai, Zeqiu Wu, Yizhong Wang, Avirup Sil, and Hannaneh Hajishirzi. 2023.
\newblock Self-rag: Learning to retrieve, generate, and critique through self-reflection.
\newblock In \emph{The Twelfth International Conference on Learning Representations}.

\bibitem[{Azaria and Mitchell(2023)}]{azaria2023internal}
Amos Azaria and Tom Mitchell. 2023.
\newblock The internal state of an llm knows when it’s lying.
\newblock In \emph{Findings of the Association for Computational Linguistics: EMNLP 2023}, pages 967--976.

\bibitem[{Baek et~al.(2023)Baek, Jeong, Kang, Park, and Hwang}]{baek2023knowledge}
Jinheon Baek, Soyeong Jeong, Minki Kang, Jong~C Park, and Sung Hwang. 2023.
\newblock Knowledge-augmented language model verification.
\newblock In \emph{Proceedings of the 2023 Conference on Empirical Methods in Natural Language Processing}, pages 1720--1736.

\bibitem[{Bansal et~al.(2023)Bansal, Samanta, Dalmia, Gupta, Ganapathy, Bapna, Jain, and Talukdar}]{Bansal2024LLMAL}
Rachit Bansal, Bidisha Samanta, Siddharth Dalmia, Nitish Gupta, Sriram Ganapathy, Abhishek Bapna, Prateek Jain, and Partha Talukdar. 2023.
\newblock Llm augmented llms: Expanding capabilities through composition.
\newblock In \emph{The Twelfth International Conference on Learning Representations}.

\bibitem[{Bashlovkina et~al.(2023)Bashlovkina, Kuang, Matthews, Clifford, Jun, Cohen, and Baumgartner}]{bashlovkina2023trusted}
Vasilisa Bashlovkina, Zhaobin Kuang, Riley Matthews, Edward Clifford, Yennie Jun, William~W Cohen, and Simon Baumgartner. 2023.
\newblock Trusted source alignment in large language models.
\newblock \emph{arXiv preprint arXiv:2311.06697}.

\bibitem[{Bender et~al.(2021)Bender, Gebru, McMillan-Major, and Shmitchell}]{bender2021dangers}
Emily~M Bender, Timnit Gebru, Angelina McMillan-Major, and Shmargaret Shmitchell. 2021.
\newblock On the dangers of stochastic parrots: Can language models be too big?
\newblock In \emph{Proceedings of the 2021 ACM conference on fairness, accountability, and transparency}, pages 610--623.

\bibitem[{Blodgett et~al.(2020)Blodgett, Barocas, Daum{\'e}~III, and Wallach}]{blodgett-etal-2020-language}
Su~Lin Blodgett, Solon Barocas, Hal Daum{\'e}~III, and Hanna Wallach. 2020.
\newblock Language (technology) is power: A critical survey of {``}bias{''} in {NLP}.
\newblock In \emph{Proceedings of the 58th Annual Meeting of the Association for Computational Linguistics}, pages 5454--5476, Online. Association for Computational Linguistics.

\bibitem[{Borgeaud et~al.(2022)Borgeaud, Mensch, Hoffmann, Cai, Rutherford, Millican, Van Den~Driessche, Lespiau, Damoc, Clark, De~Las~Casas, Guy, Menick, Ring, Hennigan, Huang, Maggiore, Jones, Cassirer, Brock, Paganini, Irving, Vinyals, Osindero, Simonyan, Rae, Elsen, and Sifre}]{borgeaud2022improving}
Sebastian Borgeaud, Arthur Mensch, Jordan Hoffmann, Trevor Cai, Eliza Rutherford, Katie Millican, George~Bm Van Den~Driessche, Jean-Baptiste Lespiau, Bogdan Damoc, Aidan Clark, Diego De~Las~Casas, Aurelia Guy, Jacob Menick, Roman Ring, Tom Hennigan, Saffron Huang, Loren Maggiore, Chris Jones, Albin Cassirer, Andy Brock, Michela Paganini, Geoffrey Irving, Oriol Vinyals, Simon Osindero, Karen Simonyan, Jack Rae, Erich Elsen, and Laurent Sifre. 2022.
\newblock Improving language models by retrieving from trillions of tokens.
\newblock In \emph{Proceedings of the 39th International Conference on Machine Learning}, volume 162 of \emph{Proceedings of Machine Learning Research}, pages 2206--2240. PMLR.

\bibitem[{Brown et~al.(2020)Brown, Mann, Ryder, Subbiah, Kaplan, Dhariwal, Neelakantan, Shyam, Sastry, Askell, Agarwal, Herbert-Voss, Krueger, Henighan, Child, Ramesh, Ziegler, Wu, Winter, Hesse, Chen, Sigler, Litwin, Gray, Chess, Clark, Berner, McCandlish, Radford, Sutskever, and Amodei}]{brown2020language}
Tom Brown, Benjamin Mann, Nick Ryder, Melanie Subbiah, Jared~D Kaplan, Prafulla Dhariwal, Arvind Neelakantan, Pranav Shyam, Girish Sastry, Amanda Askell, Sandhini Agarwal, Ariel Herbert-Voss, Gretchen Krueger, Tom Henighan, Rewon Child, Aditya Ramesh, Daniel Ziegler, Jeffrey Wu, Clemens Winter, Chris Hesse, Mark Chen, Eric Sigler, Mateusz Litwin, Scott Gray, Benjamin Chess, Jack Clark, Christopher Berner, Sam McCandlish, Alec Radford, Ilya Sutskever, and Dario Amodei. 2020.
\newblock \href {https://proceedings.neurips.cc/paper_files/paper/2020/file/1457c0d6bfcb4967418bfb8ac142f64a-Paper.pdf} {Language models are few-shot learners}.
\newblock In \emph{Advances in Neural Information Processing Systems}, volume~33, pages 1877--1901. Curran Associates, Inc.

\bibitem[{CH-Wang et~al.(2023)CH-Wang, Van~Durme, Eisner, and Kedzie}]{ch2023androids}
Sky CH-Wang, Benjamin Van~Durme, Jason Eisner, and Chris Kedzie. 2023.
\newblock Do androids know they're only dreaming of electric sheep?
\newblock \emph{arXiv preprint arXiv:2312.17249}.

\bibitem[{Chen et~al.(2023)Chen, Wang, Chen, Yu, Ma, Zhao, Yu, and Zhang}]{chen2023dense}
Tong Chen, Hongwei Wang, Sihao Chen, Wenhao Yu, Kaixin Ma, Xinran Zhao, Dong Yu, and Hongming Zhang. 2023.
\newblock Dense x retrieval: What retrieval granularity should we use?
\newblock \emph{arXiv preprint arXiv:2312.06648}.

\bibitem[{Cobbe et~al.(2021)Cobbe, Kosaraju, Bavarian, Chen, Jun, Kaiser, Plappert, Tworek, Hilton, Nakano, Hesse, and Schulman}]{cobbe2021training}
Karl Cobbe, Vineet Kosaraju, Mohammad Bavarian, Mark Chen, Heewoo Jun, Lukasz Kaiser, Matthias Plappert, Jerry Tworek, Jacob Hilton, Reiichiro Nakano, Christopher Hesse, and John Schulman. 2021.
\newblock Training verifiers to solve math word problems.
\newblock \emph{arXiv preprint arXiv:2110.14168}.

\bibitem[{Cohen et~al.(2023)Cohen, Geva, Berant, and Globerson}]{cohen2023crawling}
Roi Cohen, Mor Geva, Jonathan Berant, and Amir Globerson. 2023.
\newblock Crawling the internal knowledge-base of language models.
\newblock In \emph{Findings of the Association for Computational Linguistics: EACL 2023}, pages 1811--1824.

\bibitem[{De~Cao et~al.(2021)De~Cao, Aziz, and Titov}]{de2021editing}
Nicola De~Cao, Wilker Aziz, and Ivan Titov. 2021.
\newblock Editing factual knowledge in language models.
\newblock In \emph{Proceedings of the 2021 Conference on Empirical Methods in Natural Language Processing}, pages 6491--6506.

\bibitem[{Desai and Durrett(2020)}]{desai-durrett-2020-calibration}
Shrey Desai and Greg Durrett. 2020.
\newblock Calibration of pre-trained transformers.
\newblock In \emph{Proceedings of the 2020 Conference on Empirical Methods in Natural Language Processing (EMNLP)}, pages 295--302, Online. Association for Computational Linguistics.

\bibitem[{Ding et~al.(2023)Ding, Feng, Liu, Tan, Balachandran, He, and Tsvetkov}]{ding2023knowledge}
Wenxuan Ding, Shangbin Feng, Yuhan Liu, Zhaoxuan Tan, Vidhisha Balachandran, Tianxing He, and Yulia Tsvetkov. 2023.
\newblock Knowledge crosswords: Geometric reasoning over structured knowledge with large language models.
\newblock \emph{arXiv preprint arXiv:2310.01290}.

\bibitem[{Du et~al.(2023)Du, Li, Torralba, Tenenbaum, and Mordatch}]{du2023improving}
Yilun Du, Shuang Li, Antonio Torralba, Joshua~B Tenenbaum, and Igor Mordatch. 2023.
\newblock Improving factuality and reasoning in language models through multiagent debate.
\newblock \emph{arXiv preprint arXiv:2305.14325}.

\bibitem[{Feng et~al.(2023{\natexlab{a}})Feng, Park, Liu, and Tsvetkov}]{feng-etal-2023-pretraining}
Shangbin Feng, Chan~Young Park, Yuhan Liu, and Yulia Tsvetkov. 2023{\natexlab{a}}.
\newblock From pretraining data to language models to downstream tasks: Tracking the trails of political biases leading to unfair {NLP} models.
\newblock In \emph{Proceedings of the 61st Annual Meeting of the Association for Computational Linguistics (Volume 1: Long Papers)}, pages 11737--11762, Toronto, Canada. Association for Computational Linguistics.

\bibitem[{Feng et~al.(2023{\natexlab{b}})Feng, Shi, Bai, Balachandran, He, and Tsvetkov}]{Feng2023KnowledgeCF}
Shangbin Feng, Weijia Shi, Yuyang Bai, Vidhisha Balachandran, Tianxing He, and Yulia Tsvetkov. 2023{\natexlab{b}}.
\newblock Knowledge card: Filling llms' knowledge gaps with plug-in specialized language models.
\newblock In \emph{The Twelfth International Conference on Learning Representations}.

\bibitem[{Fu et~al.(2023)Fu, Zhang, Kwon, Perera, Zhu, Zhang, Li, Wang, Wang, Castelli, Ng, Roth, and Xiang}]{fu2023generate}
Xingyu Fu, Sheng Zhang, Gukyeong Kwon, Pramuditha Perera, Henghui Zhu, Yuhao Zhang, Alexander~Hanbo Li, William~Yang Wang, Zhiguo Wang, Vittorio Castelli, Patrick Ng, Dan Roth, and Bing Xiang. 2023.
\newblock Generate then select: Open-ended visual question answering guided by world knowledge.
\newblock In \emph{Findings of the Association for Computational Linguistics: ACL 2023}, pages 2333--2346, Toronto, Canada. Association for Computational Linguistics.

\bibitem[{Guo et~al.(2017)Guo, Pleiss, Sun, and Weinberger}]{guo2017calibration}
Chuan Guo, Geoff Pleiss, Yu~Sun, and Kilian~Q Weinberger. 2017.
\newblock On calibration of modern neural networks.
\newblock In \emph{International conference on machine learning}, pages 1321--1330. PMLR.

\bibitem[{Guu et~al.(2020)Guu, Lee, Tung, Pasupat, and Chang}]{guu2020retrieval}
Kelvin Guu, Kenton Lee, Zora Tung, Panupong Pasupat, and Mingwei Chang. 2020.
\newblock Retrieval augmented language model pre-training.
\newblock In \emph{International conference on machine learning}, pages 3929--3938. PMLR.

\bibitem[{He et~al.(2023)He, Cui, Chen, Hu, and Zhu}]{he2023investigating}
Guande He, Peng Cui, Jianfei Chen, Wenbo Hu, and Jun Zhu. 2023.
\newblock Investigating uncertainty calibration of aligned language models under the multiple-choice setting.
\newblock \emph{arXiv preprint arXiv:2310.11732}.

\bibitem[{Hendrycks et~al.(2020)Hendrycks, Burns, Basart, Zou, Mazeika, Song, and Steinhardt}]{hendrycks2020measuring}
Dan Hendrycks, Collin Burns, Steven Basart, Andy Zou, Mantas Mazeika, Dawn Song, and Jacob Steinhardt. 2020.
\newblock Measuring massive multitask language understanding.
\newblock In \emph{International Conference on Learning Representations}.

\bibitem[{Hernandez et~al.(2023)Hernandez, Li, and Andreas}]{hernandez2023measuring}
Evan Hernandez, Belinda~Z Li, and Jacob Andreas. 2023.
\newblock Measuring and manipulating knowledge representations in language models.
\newblock \emph{arXiv preprint arXiv:2304.00740}.

\bibitem[{Hu et~al.(2023)Hu, Luo, Wang, Cheng, Liu, and Sun}]{hu2023won}
Shengding Hu, Yifan Luo, Huadong Wang, Xingyi Cheng, Zhiyuan Liu, and Maosong Sun. 2023.
\newblock Won{'}t get fooled again: Answering questions with false premises.
\newblock In \emph{Proceedings of the 61st Annual Meeting of the Association for Computational Linguistics (Volume 1: Long Papers)}, pages 5626--5643, Toronto, Canada. Association for Computational Linguistics.

\bibitem[{Huang et~al.(2023{\natexlab{a}})Huang, Chen, Mishra, Zheng, Yu, Song, and Zhou}]{huang2023large}
Jie Huang, Xinyun Chen, Swaroop Mishra, Huaixiu~Steven Zheng, Adams~Wei Yu, Xinying Song, and Denny Zhou. 2023{\natexlab{a}}.
\newblock Large language models cannot self-correct reasoning yet.
\newblock In \emph{The Twelfth International Conference on Learning Representations}.

\bibitem[{Huang et~al.(2023{\natexlab{b}})Huang, Ruan, Huang, Jin, Dong, Wu, Bensalem, Mu, Qi, Zhao, Cai, Zhang, Wu, Xu, Wu, Freitas, and Mustafa}]{huang2023survey}
Xiaowei Huang, Wenjie Ruan, Wei Huang, Gao Jin, Yizhen Dong, Changshun Wu, Saddek Bensalem, Ronghui Mu, Yi~Qi, Xingyu Zhao, Kaiwen Cai, Yanghao Zhang, Sihao Wu, Peipei Xu, Dengyu Wu, Andr{\'e} Freitas, and Mustafa~A. Mustafa. 2023{\natexlab{b}}.
\newblock A survey of safety and trustworthiness of large language models through the lens of verification and validation.
\newblock \emph{ArXiv}, abs/2305.11391.

\bibitem[{Jagannatha and Yu(2020)}]{jagannatha-yu-2020-calibrating}
Abhyuday Jagannatha and Hong Yu. 2020.
\newblock \href {https://doi.org/10.18653/v1/2020.acl-main.188} {Calibrating structured output predictors for natural language processing}.
\newblock In \emph{Proceedings of the 58th Annual Meeting of the Association for Computational Linguistics}, pages 2078--2092, Online. Association for Computational Linguistics.

\bibitem[{Ji et~al.(2023)Ji, Lee, Frieske, Yu, Su, Xu, Ishii, Bang, Madotto, and Fung}]{ji2023survey}
Ziwei Ji, Nayeon Lee, Rita Frieske, Tiezheng Yu, Dan Su, Yan Xu, Etsuko Ishii, Ye~Jin Bang, Andrea Madotto, and Pascale Fung. 2023.
\newblock Survey of hallucination in natural language generation.
\newblock \emph{ACM Computing Surveys}, 55(12):1--38.

\bibitem[{Jiang et~al.(2023)Jiang, Sablayrolles, Mensch, Bamford, Chaplot, de~Las~Casas, Bressand, Lengyel, Lample, Saulnier, Lavaud, Lachaux, Stock, Scao, Lavril, Wang, Lacroix, and Sayed}]{jiang2023mistral}
Albert~Qiaochu Jiang, Alexandre Sablayrolles, Arthur Mensch, Chris Bamford, Devendra~Singh Chaplot, Diego de~Las~Casas, Florian Bressand, Gianna Lengyel, Guillaume Lample, Lucile Saulnier, L'elio~Renard Lavaud, Marie-Anne Lachaux, Pierre Stock, Teven~Le Scao, Thibaut Lavril, Thomas Wang, Timoth{\'e}e Lacroix, and William~El Sayed. 2023.
\newblock Mistral 7b.
\newblock \emph{ArXiv}, abs/2310.06825.

\bibitem[{Jiang et~al.(2021)Jiang, Araki, Ding, and Neubig}]{jiang2021can}
Zhengbao Jiang, Jun Araki, Haibo Ding, and Graham Neubig. 2021.
\newblock How can we know when language models know? on the calibration of language models for question answering.
\newblock \emph{Transactions of the Association for Computational Linguistics}, 9:962--977.

\bibitem[{Jin et~al.(2021)Jin, Barbieri, Kennedy, Mostafazadeh~Davani, Neves, and Ren}]{jin-etal-2021-transferability}
Xisen Jin, Francesco Barbieri, Brendan Kennedy, Aida Mostafazadeh~Davani, Leonardo Neves, and Xiang Ren. 2021.
\newblock On transferability of bias mitigation effects in language model fine-tuning.
\newblock In \emph{Proceedings of the 2021 Conference of the North American Chapter of the Association for Computational Linguistics: Human Language Technologies}, pages 3770--3783, Online. Association for Computational Linguistics.

\bibitem[{Kadavath et~al.(2022)Kadavath, Conerly, Askell, Henighan, Drain, Perez, Schiefer, Dodds, DasSarma, Tran-Johnson, Johnston, El-Showk, Jones, Elhage, Hume, Chen, Bai, Bowman, Fort, Ganguli, Hernandez, Jacobson, Kernion, Kravec, Lovitt, Ndousse, Olsson, Ringer, Amodei, Brown, Clark, Joseph, Mann, McCandlish, Olah, and Kaplan}]{kadavath2022language}
Saurav Kadavath, Tom Conerly, Amanda Askell, T.~J. Henighan, Dawn Drain, Ethan Perez, Nicholas Schiefer, Zachary Dodds, Nova DasSarma, Eli Tran-Johnson, Scott Johnston, Sheer El-Showk, Andy Jones, Nelson Elhage, Tristan Hume, Anna Chen, Yuntao Bai, Sam Bowman, Stanislav Fort, Deep Ganguli, Danny Hernandez, Josh Jacobson, John Kernion, Shauna Kravec, Liane Lovitt, Kamal Ndousse, Catherine Olsson, Sam Ringer, Dario Amodei, Tom~B. Brown, Jack Clark, Nicholas Joseph, Benjamin Mann, Sam McCandlish, Christopher Olah, and Jared Kaplan. 2022.
\newblock Language models (mostly) know what they know.
\newblock \emph{ArXiv}, abs/2207.05221.

\bibitem[{Kalai and Vempala(2023)}]{kalai2023calibrated}
Adam~Tauman Kalai and Santosh~S Vempala. 2023.
\newblock Calibrated language models must hallucinate.
\newblock \emph{arXiv preprint arXiv:2311.14648}.

\bibitem[{Kamath et~al.(2020)Kamath, Jia, and Liang}]{kamath-etal-2020-selective}
Amita Kamath, Robin Jia, and Percy Liang. 2020.
\newblock Selective question answering under domain shift.
\newblock In \emph{Proceedings of the 58th Annual Meeting of the Association for Computational Linguistics}, pages 5684--5696, Online. Association for Computational Linguistics.

\bibitem[{Kandpal et~al.(2023)Kandpal, Deng, Roberts, Wallace, and Raffel}]{kandpal2023large}
Nikhil Kandpal, Haikang Deng, Adam Roberts, Eric Wallace, and Colin Raffel. 2023.
\newblock Large language models struggle to learn long-tail knowledge.
\newblock In \emph{International Conference on Machine Learning}, pages 15696--15707. PMLR.

\bibitem[{Kang et~al.(2023)Kang, Luo, Zhu, Glass, Cox, Ritter, Feris, and Karlinsky}]{kang2023self}
Junmo Kang, Hongyin Luo, Yada Zhu, James Glass, David Cox, Alan Ritter, Rogerio Feris, and Leonid Karlinsky. 2023.
\newblock Self-specialization: Uncovering latent expertise within large language models.
\newblock \emph{arXiv preprint arXiv:2310.00160}.

\bibitem[{Kasai et~al.(2024)Kasai, Sakaguchi, Le~Bras, Asai, Yu, Radev, Smith, Choi, Inui et~al.}]{kasai2022realtime}
Jungo Kasai, Keisuke Sakaguchi, Ronan Le~Bras, Akari Asai, Xinyan Yu, Dragomir Radev, Noah~A Smith, Yejin Choi, Kentaro Inui, et~al. 2024.
\newblock Realtime qa: What's the answer right now?
\newblock \emph{Advances in Neural Information Processing Systems}, 36.

\bibitem[{Khattab et~al.(2022)Khattab, Santhanam, Li, Hall, Liang, Potts, and Zaharia}]{khattab2022demonstrate}
Omar Khattab, Keshav Santhanam, Xiang~Lisa Li, David Hall, Percy Liang, Christopher Potts, and Matei Zaharia. 2022.
\newblock Demonstrate-search-predict: Composing retrieval and language models for knowledge-intensive nlp.
\newblock \emph{arXiv preprint arXiv:2212.14024}.

\bibitem[{Kim et~al.(2024)Kim, Baldi, and McAleer}]{kim2023language}
Geunwoo Kim, Pierre Baldi, and Stephen McAleer. 2024.
\newblock Language models can solve computer tasks.
\newblock \emph{Advances in Neural Information Processing Systems}, 36.

\bibitem[{Kong et~al.(2020)Kong, Jiang, Zhuang, Lyu, Zhao, and Zhang}]{kong-etal-2020-calibrated}
Lingkai Kong, Haoming Jiang, Yuchen Zhuang, Jie Lyu, Tuo Zhao, and Chao Zhang. 2020.
\newblock Calibrated language model fine-tuning for in- and out-of-distribution data.
\newblock In \emph{Proceedings of the 2020 Conference on Empirical Methods in Natural Language Processing (EMNLP)}, pages 1326--1340, Online. Association for Computational Linguistics.

\bibitem[{Kuhn et~al.(2022)Kuhn, Gal, and Farquhar}]{kuhn2023semantic}
Lorenz Kuhn, Yarin Gal, and Sebastian Farquhar. 2022.
\newblock Semantic uncertainty: Linguistic invariances for uncertainty estimation in natural language generation.
\newblock In \emph{The Eleventh International Conference on Learning Representations}.

\bibitem[{Liang et~al.(2023)Liang, Bommasani, Lee, Tsipras, Soylu, Yasunaga, Zhang, Narayanan, Wu, Kumar et~al.}]{liang2022holistic}
Percy Liang, Rishi Bommasani, Tony Lee, Dimitris Tsipras, Dilara Soylu, Michihiro Yasunaga, Yian Zhang, Deepak Narayanan, Yuhuai Wu, Ananya Kumar, et~al. 2023.
\newblock Holistic evaluation of language models.
\newblock \emph{Transactions on Machine Learning Research}.

\bibitem[{Lin et~al.(2022)Lin, Hilton, and Evans}]{lin2022teaching}
Stephanie Lin, Jacob Hilton, and Owain Evans. 2022.
\newblock Teaching models to express their uncertainty in words.
\newblock \emph{arXiv preprint arXiv:2205.14334}.

\bibitem[{Liu et~al.(2023{\natexlab{a}})Liu, Khalifa, and Wang}]{liu2023litcab}
Xin Liu, Muhammad Khalifa, and Lu~Wang. 2023{\natexlab{a}}.
\newblock Litcab: Lightweight calibration of language models on outputs of varied lengths.
\newblock \emph{arXiv preprint arXiv:2310.19208}.

\bibitem[{Liu et~al.(2023{\natexlab{b}})Liu, Yao, Ton, Zhang, Guo, Cheng, Klochkov, Taufiq, and Li}]{liu2023trustworthy}
Yang Liu, Yuanshun Yao, Jean-Francois Ton, Xiaoying Zhang, Ruocheng Guo, Hao Cheng, Yegor Klochkov, Muhammad~Faaiz Taufiq, and Hang Li. 2023{\natexlab{b}}.
\newblock Trustworthy llms: a survey and guideline for evaluating large language models' alignment.
\newblock In \emph{Socially Responsible Language Modelling Research}.

\bibitem[{Liu et~al.(2019)Liu, Ott, Goyal, Du, Joshi, Chen, Levy, Lewis, Zettlemoyer, and Stoyanov}]{liu2019roberta}
Yinhan Liu, Myle Ott, Naman Goyal, Jingfei Du, Mandar Joshi, Danqi Chen, Omer Levy, Mike Lewis, Luke Zettlemoyer, and Veselin Stoyanov. 2019.
\newblock Roberta: A robustly optimized bert pretraining approach.
\newblock \emph{arXiv preprint arXiv:1907.11692}.

\bibitem[{Luo et~al.(2023)Luo, Xu, Zhao, Geng, Tao, Ma, Lin, and Jiang}]{luo2023augmented}
Ziyang Luo, Can Xu, Pu~Zhao, Xiubo Geng, Chongyang Tao, Jing Ma, Qingwei Lin, and Daxin Jiang. 2023.
\newblock Augmented large language models with parametric knowledge guiding.
\newblock \emph{arXiv preprint arXiv:2305.04757}.

\bibitem[{Mallen et~al.(2023)Mallen, Asai, Zhong, Das, Khashabi, and Hajishirzi}]{mallen2023not}
Alex Mallen, Akari Asai, Victor Zhong, Rajarshi Das, Daniel Khashabi, and Hannaneh Hajishirzi. 2023.
\newblock When not to trust language models: Investigating effectiveness of parametric and non-parametric memories.
\newblock In \emph{Proceedings of the 61st Annual Meeting of the Association for Computational Linguistics (Volume 1: Long Papers)}, pages 9802--9822.

\bibitem[{Mielke et~al.(2022)Mielke, Szlam, Dinan, and Boureau}]{mielke2022reducing}
Sabrina~J Mielke, Arthur Szlam, Emily Dinan, and Y-Lan Boureau. 2022.
\newblock Reducing conversational agents’ overconfidence through linguistic calibration.
\newblock \emph{Transactions of the Association for Computational Linguistics}, 10:857--872.

\bibitem[{Min et~al.(2020)Min, Michael, Hajishirzi, and Zettlemoyer}]{min2020ambigqa}
Sewon Min, Julian Michael, Hannaneh Hajishirzi, and Luke Zettlemoyer. 2020.
\newblock {A}mbig{QA}: Answering ambiguous open-domain questions.
\newblock In \emph{Proceedings of the 2020 Conference on Empirical Methods in Natural Language Processing (EMNLP)}.

\bibitem[{Nakano et~al.(2021)Nakano, Hilton, Balaji, Wu, Long, Kim, Hesse, Jain, Kosaraju, Saunders, Jiang, Cobbe, Eloundou, Krueger, Button, Knight, Chess, and Schulman}]{nakano2021webgpt}
Reiichiro Nakano, Jacob Hilton, Suchir Balaji, Jeff Wu, Ouyang Long, Christina Kim, Christopher Hesse, Shantanu Jain, Vineet Kosaraju, William Saunders, Xu~Jiang, Karl Cobbe, Tyna Eloundou, Gretchen Krueger, Kevin Button, Matthew Knight, Benjamin Chess, and John Schulman. 2021.
\newblock Webgpt: Browser-assisted question-answering with human feedback.
\newblock \emph{ArXiv}, abs/2112.09332.

\bibitem[{Ouyang et~al.(2022)Ouyang, Wu, Jiang, Almeida, Wainwright, Mishkin, Zhang, Agarwal, Slama, Ray et~al.}]{ouyang2022training}
Long Ouyang, Jeffrey Wu, Xu~Jiang, Diogo Almeida, Carroll Wainwright, Pamela Mishkin, Chong Zhang, Sandhini Agarwal, Katarina Slama, Alex Ray, et~al. 2022.
\newblock Training language models to follow instructions with human feedback.
\newblock \emph{Advances in Neural Information Processing Systems}, 35:27730--27744.

\bibitem[{Patidar et~al.(2023)Patidar, Faldu, Singh, Vig, Bhattacharya, and {Mausam}}]{patidar2022have}
Mayur Patidar, Prayushi Faldu, Avinash Singh, Lovekesh Vig, Indrajit Bhattacharya, and {Mausam}. 2023.
\newblock Do {I} have the knowledge to answer? investigating answerability of knowledge base questions.
\newblock In \emph{Proceedings of the 61st Annual Meeting of the Association for Computational Linguistics (Volume 1: Long Papers)}, pages 10341--10357, Toronto, Canada. Association for Computational Linguistics.

\bibitem[{Petroni et~al.(2019)Petroni, Rockt{\"a}schel, Riedel, Lewis, Bakhtin, Wu, and Miller}]{petroni2019language}
Fabio Petroni, Tim Rockt{\"a}schel, Sebastian Riedel, Patrick Lewis, Anton Bakhtin, Yuxiang Wu, and Alexander Miller. 2019.
\newblock Language models as knowledge bases?
\newblock In \emph{Proceedings of the 2019 Conference on Empirical Methods in Natural Language Processing and the 9th International Joint Conference on Natural Language Processing (EMNLP-IJCNLP)}, pages 2463--2473.

\bibitem[{Piskorski et~al.(2023)Piskorski, Stefanovitch, Da~San~Martino, and Nakov}]{piskorski2023semeval}
Jakub Piskorski, Nicolas Stefanovitch, Giovanni Da~San~Martino, and Preslav Nakov. 2023.
\newblock Semeval-2023 task 3: Detecting the category, the framing, and the persuasion techniques in online news in a multi-lingual setup.
\newblock In \emph{Proceedings of the the 17th International Workshop on Semantic Evaluation (SemEval-2023)}, pages 2343--2361.

\bibitem[{Press et~al.(2023)Press, Zhang, Min, Schmidt, Smith, and Lewis}]{press-etal-2023-measuring}
Ofir Press, Muru Zhang, Sewon Min, Ludwig Schmidt, Noah Smith, and Mike Lewis. 2023.
\newblock Measuring and narrowing the compositionality gap in language models.
\newblock In \emph{Findings of the Association for Computational Linguistics: EMNLP 2023}, pages 5687--5711, Singapore. Association for Computational Linguistics.

\bibitem[{Radford et~al.(2019)Radford, Wu, Child, Luan, Amodei, Sutskever et~al.}]{radford2019language}
Alec Radford, Jeffrey Wu, Rewon Child, David Luan, Dario Amodei, Ilya Sutskever, et~al. 2019.
\newblock Language models are unsupervised multitask learners.
\newblock \emph{OpenAI blog}, 1(8):9.

\bibitem[{Sclar et~al.(2023)Sclar, Choi, Tsvetkov, and Suhr}]{sclar2023quantifying}
Melanie Sclar, Yejin Choi, Yulia Tsvetkov, and Alane Suhr. 2023.
\newblock Quantifying language models' sensitivity to spurious features in prompt design or: How i learned to start worrying about prompt formatting.
\newblock In \emph{The Twelfth International Conference on Learning Representations}.

\bibitem[{Shaikh et~al.(2023)Shaikh, Zhang, Held, Bernstein, and Yang}]{shaikh-etal-2023-second}
Omar Shaikh, Hongxin Zhang, William Held, Michael Bernstein, and Diyi Yang. 2023.
\newblock On second thought, let{'}s not think step by step! bias and toxicity in zero-shot reasoning.
\newblock In \emph{Proceedings of the 61st Annual Meeting of the Association for Computational Linguistics (Volume 1: Long Papers)}, pages 4454--4470, Toronto, Canada. Association for Computational Linguistics.

\bibitem[{Shi et~al.(2023)Shi, Min, Yasunaga, Seo, James, Lewis, Zettlemoyer, and Yih}]{shi2023replug}
Weijia Shi, Sewon Min, Michihiro Yasunaga, Minjoon Seo, Rich James, Mike Lewis, Luke Zettlemoyer, and Wen-tau Yih. 2023.
\newblock Replug: Retrieval-augmented black-box language models.
\newblock \emph{arXiv preprint arXiv:2301.12652}.

\bibitem[{Shinn et~al.(2023)Shinn, Cassano, Gopinath, Narasimhan, and Yao}]{shinn2023reflexion}
Noah Shinn, Federico Cassano, Ashwin Gopinath, Karthik~R Narasimhan, and Shunyu Yao. 2023.
\newblock Reflexion: Language agents with verbal reinforcement learning.
\newblock In \emph{Thirty-seventh Conference on Neural Information Processing Systems}.

\bibitem[{Si et~al.(2023)Si, Shi, Zhao, Zettlemoyer, and Boyd-Graber}]{si2023getting}
Chenglei Si, Weijia Shi, Chen Zhao, Luke Zettlemoyer, and Jordan Boyd-Graber. 2023.
\newblock Getting more out of mixture of language model reasoning experts.
\newblock In \emph{Findings of the Association for Computational Linguistics: EMNLP 2023}, pages 8234--8249.

\bibitem[{Slobodkin et~al.(2023)Slobodkin, Goldman, Caciularu, Dagan, and Ravfogel}]{slobodkin2023curious}
Aviv Slobodkin, Omer Goldman, Avi Caciularu, Ido Dagan, and Shauli Ravfogel. 2023.
\newblock The curious case of hallucinatory (un) answerability: Finding truths in the hidden states of over-confident large language models.
\newblock In \emph{Proceedings of the 2023 Conference on Empirical Methods in Natural Language Processing}, pages 3607--3625.

\bibitem[{Stengel-Eskin and Van~Durme(2023)}]{stengel2023calibrated}
Elias Stengel-Eskin and Benjamin Van~Durme. 2023.
\newblock Calibrated interpretation: Confidence estimation in semantic parsing.
\newblock \emph{Transactions of the Association for Computational Linguistics}, 11:1213--1231.

\bibitem[{Sun et~al.(2022)Sun, Yan, Abbeel, and Mordatch}]{sun2022quantifying}
Meiqi Sun, Wilson Yan, Pieter Abbeel, and Igor Mordatch. 2022.
\newblock Quantifying uncertainty in foundation models via ensembles.
\newblock In \emph{NeurIPS 2022 Workshop on Robustness in Sequence Modeling}.

\bibitem[{Sun et~al.(2023)Sun, Shen, Cao, Liu, Li, Shen, Gan, Gui, Wang, Yang et~al.}]{sun2023aligning}
Zhiqing Sun, Sheng Shen, Shengcao Cao, Haotian Liu, Chunyuan Li, Yikang Shen, Chuang Gan, Liang-Yan Gui, Yu-Xiong Wang, Yiming Yang, et~al. 2023.
\newblock Aligning large multimodal models with factually augmented rlhf.
\newblock \emph{arXiv preprint arXiv:2309.14525}.

\bibitem[{Tao et~al.(2023)Tao, Zhu, Guo, Dong, and Xu}]{tao2023benchmark}
Linwei Tao, Younan Zhu, Haolan Guo, Minjing Dong, and Chang Xu. 2023.
\newblock A benchmark study on calibration.
\newblock In \emph{The Twelfth International Conference on Learning Representations}.

\bibitem[{Thakur et~al.(2023)Thakur, Bonifacio, Zhang, Ogundepo, Kamalloo, Alfonso-Hermelo, Li, Liu, Chen, Rezagholizadeh et~al.}]{thakur2023nomiracl}
Nandan Thakur, Luiz Bonifacio, Xinyu Zhang, Odunayo Ogundepo, Ehsan Kamalloo, David Alfonso-Hermelo, Xiaoguang Li, Qun Liu, Boxing Chen, Mehdi Rezagholizadeh, et~al. 2023.
\newblock Nomiracl: Knowing when you don't know for robust multilingual retrieval-augmented generation.
\newblock \emph{arXiv preprint arXiv:2312.11361}.

\bibitem[{Tian et~al.(2023)Tian, Mitchell, Zhou, Sharma, Rafailov, Yao, Finn, and Manning}]{tian2023just}
Katherine Tian, Eric Mitchell, Allan Zhou, Archit Sharma, Rafael Rafailov, Huaxiu Yao, Chelsea Finn, and Christopher Manning. 2023.
\newblock Just ask for calibration: Strategies for eliciting calibrated confidence scores from language models fine-tuned with human feedback.
\newblock In \emph{Proceedings of the 2023 Conference on Empirical Methods in Natural Language Processing}, pages 5433--5442, Singapore. Association for Computational Linguistics.

\bibitem[{Touvron et~al.(2023)Touvron, Martin, Stone, Albert, Almahairi, Babaei, Bashlykov, Batra, Bhargava, Bhosale et~al.}]{touvron2023llama}
Hugo Touvron, Louis Martin, Kevin Stone, Peter Albert, Amjad Almahairi, Yasmine Babaei, Nikolay Bashlykov, Soumya Batra, Prajjwal Bhargava, Shruti Bhosale, et~al. 2023.
\newblock Llama 2: Open foundation and fine-tuned chat models.
\newblock \emph{arXiv preprint arXiv:2307.09288}.

\bibitem[{Wang et~al.(2023{\natexlab{a}})Wang, Yue, and Sun}]{wang-etal-2023-chatgpt-defend}
Boshi Wang, Xiang Yue, and Huan Sun. 2023{\natexlab{a}}.
\newblock Can {C}hat{GPT} defend its belief in truth? evaluating {LLM} reasoning via debate.
\newblock In \emph{Findings of the Association for Computational Linguistics: EMNLP 2023}, pages 11865--11881, Singapore. Association for Computational Linguistics.

\bibitem[{Wang et~al.(2020)Wang, Tu, Shi, and Liu}]{wang-etal-2020-inference}
Shuo Wang, Zhaopeng Tu, Shuming Shi, and Yang Liu. 2020.
\newblock On the inference calibration of neural machine translation.
\newblock In \emph{Proceedings of the 58th Annual Meeting of the Association for Computational Linguistics}, pages 3070--3079, Online. Association for Computational Linguistics.

\bibitem[{Wang et~al.(2022)Wang, Wei, Schuurmans, Le, Chi, Narang, Chowdhery, and Zhou}]{wang2022self}
Xuezhi Wang, Jason Wei, Dale Schuurmans, Quoc~V Le, Ed~H Chi, Sharan Narang, Aakanksha Chowdhery, and Denny Zhou. 2022.
\newblock Self-consistency improves chain of thought reasoning in language models.
\newblock In \emph{The Eleventh International Conference on Learning Representations}.

\bibitem[{Wang et~al.(2023{\natexlab{b}})Wang, Feng, Wang, Shi, Balachandran, He, and Tsvetkov}]{wang2023resolving}
Yike Wang, Shangbin Feng, Heng Wang, Weijia Shi, Vidhisha Balachandran, Tianxing He, and Yulia Tsvetkov. 2023{\natexlab{b}}.
\newblock Resolving knowledge conflicts in large language models.
\newblock \emph{arXiv preprint arXiv:2310.00935}.

\bibitem[{Wang et~al.(2023{\natexlab{c}})Wang, Araki, Jiang, Parvez, and Neubig}]{wang2023learning}
Zhiruo Wang, Jun Araki, Zhengbao Jiang, Md~Rizwan Parvez, and Graham Neubig. 2023{\natexlab{c}}.
\newblock Learning to filter context for retrieval-augmented generation.
\newblock \emph{arXiv preprint arXiv:2311.08377}.

\bibitem[{Wei et~al.(2022)Wei, Wang, Schuurmans, Bosma, Xia, Chi, Le, Zhou et~al.}]{wei2022chain}
Jason Wei, Xuezhi Wang, Dale Schuurmans, Maarten Bosma, Fei Xia, Ed~Chi, Quoc~V Le, Denny Zhou, et~al. 2022.
\newblock Chain-of-thought prompting elicits reasoning in large language models.
\newblock \emph{Advances in Neural Information Processing Systems}, 35:24824--24837.

\bibitem[{West et~al.(2023)West, Lu, Dziri, Brahman, Li, Hwang, Jiang, Fisher, Ravichander, Chandu et~al.}]{west2023generative}
Peter West, Ximing Lu, Nouha Dziri, Faeze Brahman, Linjie Li, Jena~D Hwang, Liwei Jiang, Jillian Fisher, Abhilasha Ravichander, Khyathi Chandu, et~al. 2023.
\newblock The generative ai paradox:“what it can create, it may not understand”.
\newblock In \emph{The Twelfth International Conference on Learning Representations}.

\bibitem[{Whitehead et~al.(2022)Whitehead, Petryk, Shakib, Gonzalez, Darrell, Rohrbach, and Rohrbach}]{whitehead2022reliable}
Spencer Whitehead, Suzanne Petryk, Vedaad Shakib, Joseph Gonzalez, Trevor Darrell, Anna Rohrbach, and Marcus Rohrbach. 2022.
\newblock Reliable visual question answering: Abstain rather than answer incorrectly.
\newblock In \emph{European Conference on Computer Vision}, pages 148--166. Springer.

\bibitem[{Wolf et~al.(2019)Wolf, Debut, Sanh, Chaumond, Delangue, Moi, Cistac, Rault, Louf, Funtowicz, and Brew}]{wolf2019huggingface}
Thomas Wolf, Lysandre Debut, Victor Sanh, Julien Chaumond, Clement Delangue, Anthony Moi, Pierric Cistac, Tim Rault, R{\'e}mi Louf, Morgan Funtowicz, and Jamie Brew. 2019.
\newblock Huggingface's transformers: State-of-the-art natural language processing.
\newblock \emph{ArXiv}, abs/1910.03771.

\bibitem[{Xie et~al.(2023)Xie, Zhang, Chen, Lou, and Su}]{xie2023adaptive}
Jian Xie, Kai Zhang, Jiangjie Chen, Renze Lou, and Yu~Su. 2023.
\newblock Adaptive chameleon or stubborn sloth: Revealing the behavior of large language models in knowledge conflicts.
\newblock In \emph{The Twelfth International Conference on Learning Representations}.

\bibitem[{Xu et~al.(2023)Xu, Shi, and Choi}]{xu2023recomp}
Fangyuan Xu, Weijia Shi, and Eunsol Choi. 2023.
\newblock Recomp: Improving retrieval-augmented lms with compression and selective augmentation.
\newblock \emph{arXiv preprint arXiv:2310.04408}.

\bibitem[{Yang et~al.(2023)Yang, Chern, Qiu, Neubig, and Liu}]{yang2023alignment}
Yuqing Yang, Ethan Chern, Xipeng Qiu, Graham Neubig, and Pengfei Liu. 2023.
\newblock Alignment for honesty.
\newblock \emph{arXiv preprint arXiv:2312.07000}.

\bibitem[{Yu et~al.(2023{\natexlab{a}})Yu, Wang, Tu, Cao, Zhang-Li, Lv, Peng, Yao, Zhang, Li et~al.}]{yu2023kola}
Jifan Yu, Xiaozhi Wang, Shangqing Tu, Shulin Cao, Daniel Zhang-Li, Xin Lv, Hao Peng, Zijun Yao, Xiaohan Zhang, Hanming Li, et~al. 2023{\natexlab{a}}.
\newblock Kola: Carefully benchmarking world knowledge of large language models.
\newblock In \emph{The Twelfth International Conference on Learning Representations}.

\bibitem[{Yu et~al.(2023{\natexlab{b}})Yu, Zhang, Pan, Ma, Wang, and Yu}]{yu2023chain}
Wenhao Yu, Hongming Zhang, Xiaoman Pan, Kaixin Ma, Hongwei Wang, and Dong Yu. 2023{\natexlab{b}}.
\newblock Chain-of-note: Enhancing robustness in retrieval-augmented language models.
\newblock \emph{arXiv preprint arXiv:2311.09210}.

\bibitem[{Zellers et~al.(2019)Zellers, Holtzman, Bisk, Farhadi, and Choi}]{zellers2019hellaswag}
Rowan Zellers, Ari Holtzman, Yonatan Bisk, Ali Farhadi, and Yejin Choi. 2019.
\newblock Hellaswag: Can a machine really finish your sentence?
\newblock In \emph{Proceedings of the 57th Annual Meeting of the Association for Computational Linguistics}, pages 4791--4800.

\bibitem[{Zhang et~al.(2023)Zhang, Diao, Lin, Fung, Lian, Wang, Chen, Ji, and Zhang}]{zhang2023r}
Hanning Zhang, Shizhe Diao, Yong Lin, Yi~R Fung, Qing Lian, Xingyao Wang, Yangyi Chen, Heng Ji, and Tong Zhang. 2023.
\newblock R-tuning: Teaching large language models to refuse unknown questions.
\newblock \emph{arXiv preprint arXiv:2311.09677}.

\bibitem[{Zhou et~al.(2023{\natexlab{a}})Zhou, Wan, Proleev, Mincu, Chen, Heller, and Roy}]{zhou2023batch}
Han Zhou, Xingchen Wan, Lev Proleev, Diana Mincu, Jilin Chen, Katherine Heller, and Subhrajit Roy. 2023{\natexlab{a}}.
\newblock Batch calibration: Rethinking calibration for in-context learning and prompt engineering.
\newblock \emph{arXiv preprint arXiv:2309.17249}.

\bibitem[{Zhou et~al.(2024)Zhou, Hwang, Ren, and Sap}]{Zhou2024RelyingOT}
Kaitlyn Zhou, Jena~D Hwang, Xiang Ren, and Maarten Sap. 2024.
\newblock Relying on the unreliable: The impact of language models' reluctance to express uncertainty.
\newblock \emph{arXiv preprint arXiv:2401.06730}.

\bibitem[{Zhou et~al.(2023{\natexlab{b}})Zhou, Jurafsky, and Hashimoto}]{zhou-etal-2023-navigating}
Kaitlyn Zhou, Dan Jurafsky, and Tatsunori Hashimoto. 2023{\natexlab{b}}.
\newblock Navigating the grey area: How expressions of uncertainty and overconfidence affect language models.
\newblock In \emph{Proceedings of the 2023 Conference on Empirical Methods in Natural Language Processing}, pages 5506--5524, Singapore. Association for Computational Linguistics.

\end{thebibliography}

\newpage

\appendix

\section{Analysis (cont.)}
\label{sec:analysis_cont}

\begin{figure}[t]
    \centering
    \includegraphics[width=\linewidth]{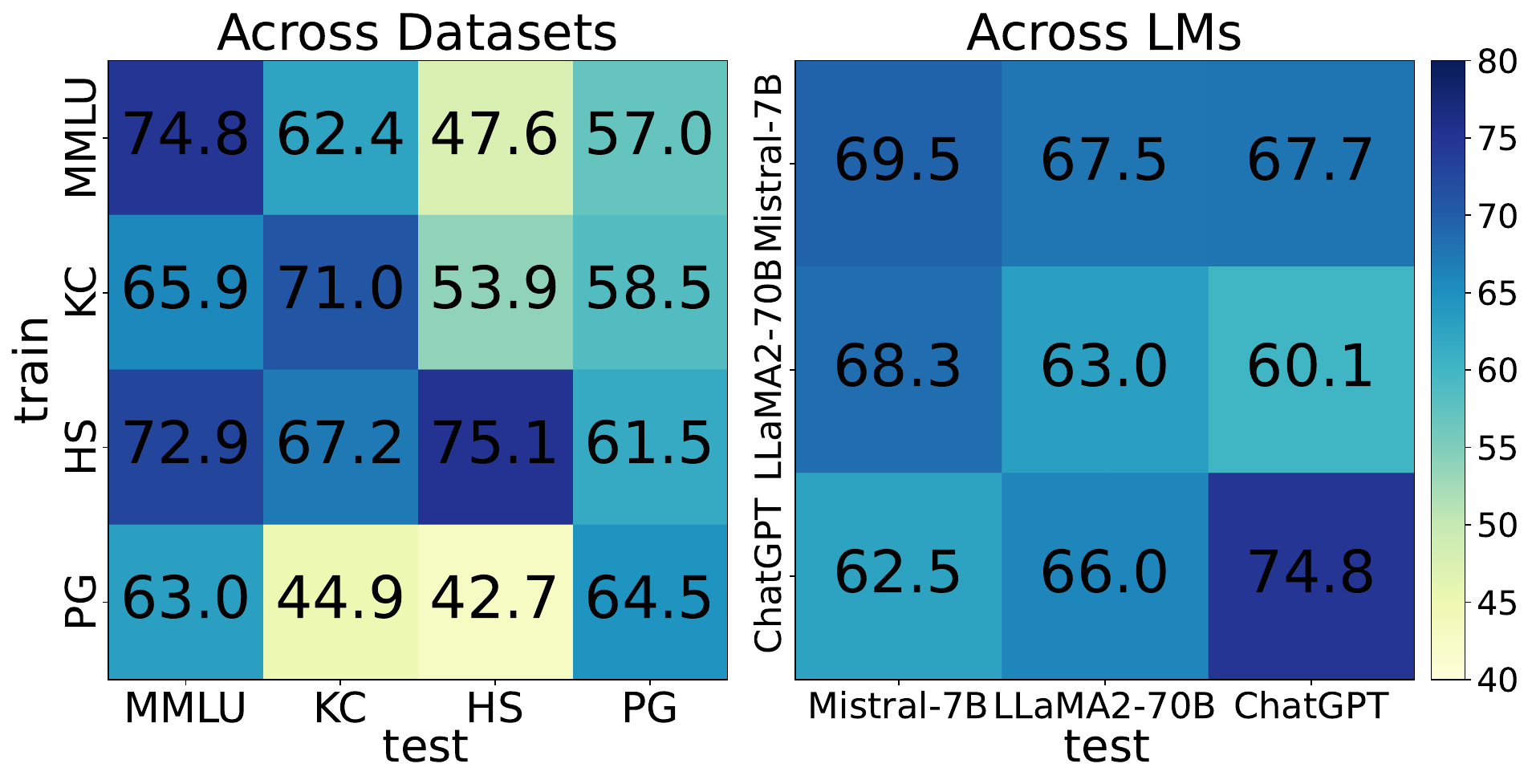}
    \caption{Performance of \textsc{instruction tuning} when trained with one dataset/LLM and tested for another. Training on a held-out set harms the generalization of abstain accuracy across tasks and LLMs.}
    \label{fig:domain transfer}
\end{figure}

\begin{figure}[t]
    \centering
    \includegraphics[width=1\linewidth]{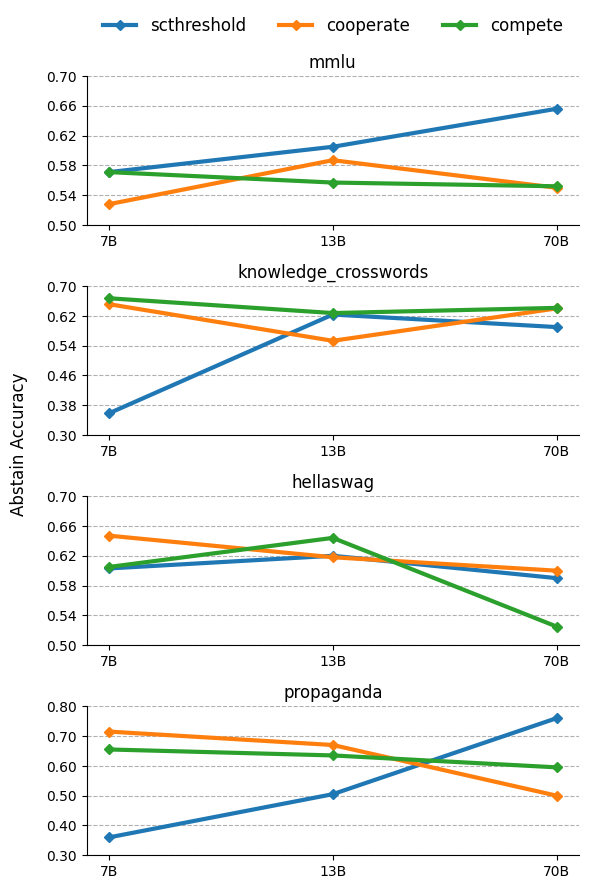}
    \caption{Performance with three abstain mechanisms with increasing size of model parameters.}
    \label{fig:scaling_acc}
\end{figure}

\begin{figure*}[t]
    \centering
    \includegraphics[width=\linewidth]{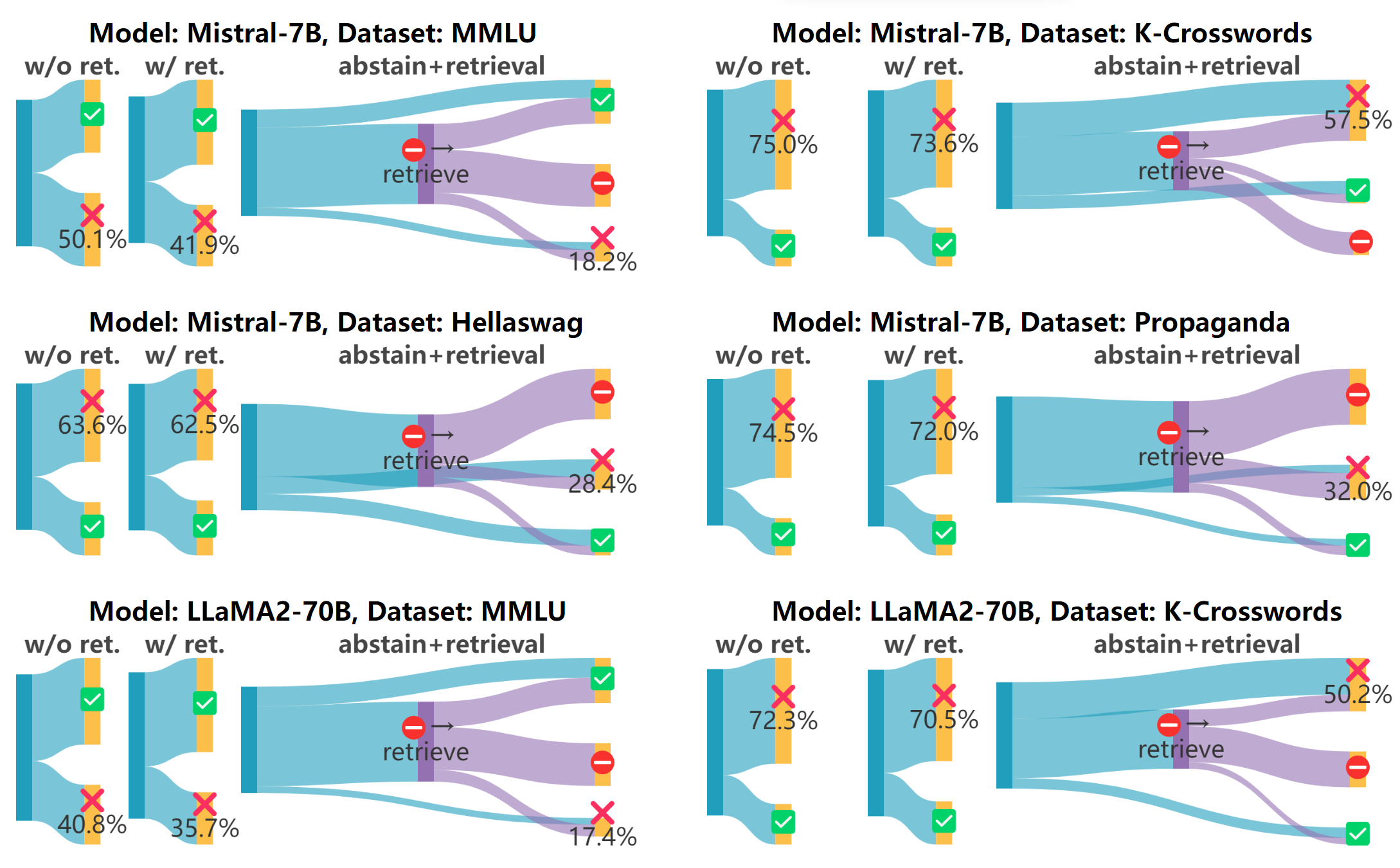}
    \includegraphics[width=\linewidth]{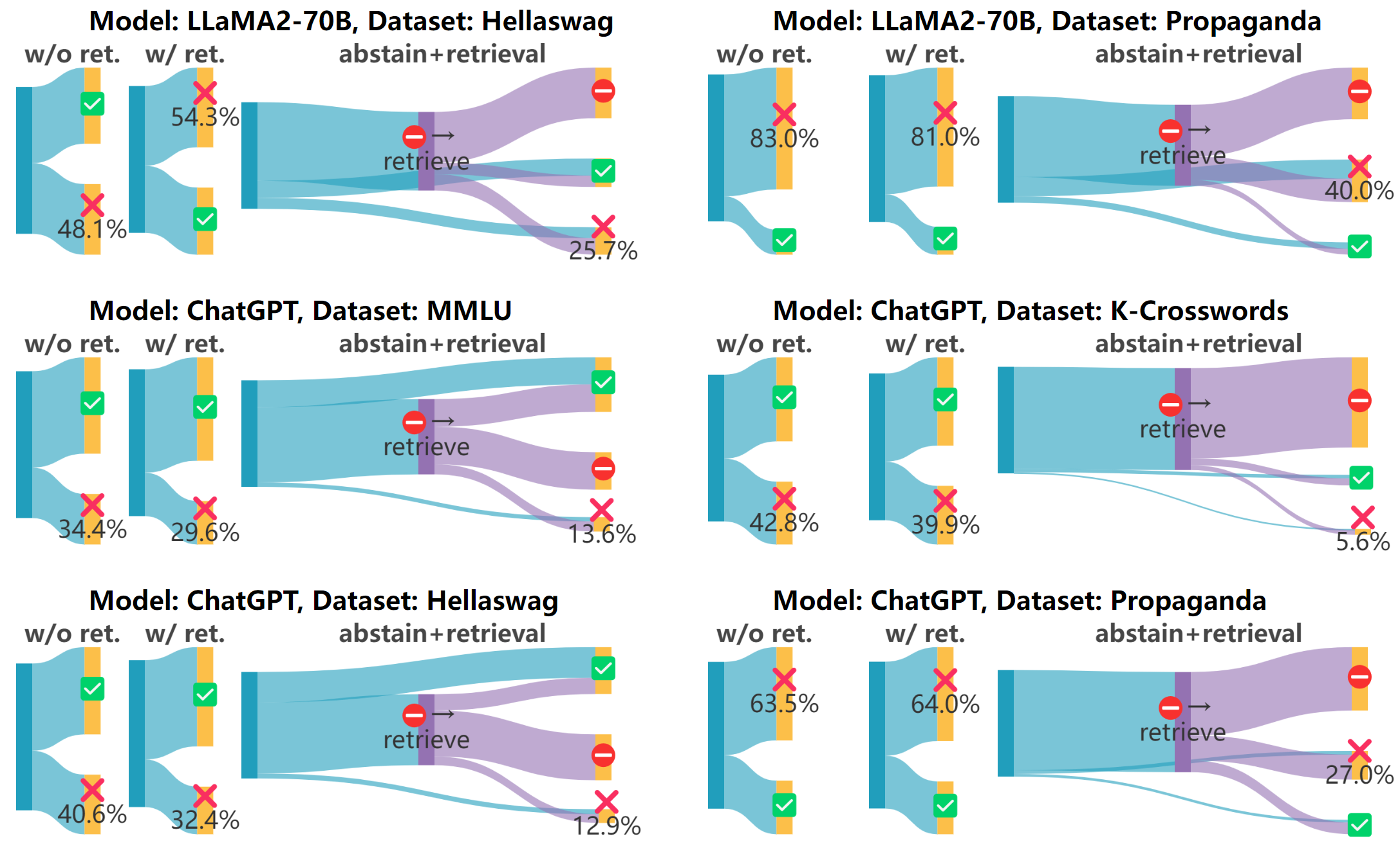}
        \caption{Model performance of \emph{without retrieval}, \emph{with retrieval}, and \emph{retrieval+abstain} settings with the \textsc{Compete} approach.}
    \label{fig:retrieval_failure_cont}
\end{figure*}

\paragraph{Retrieval Failure (cont.)} We present more results with the two-step abstention in retrieval-augmented LLMs in Figure \ref{fig:retrieval_failure_cont}. Across three LLMs and four datasets, the abstain-retrieve-abstain pipeline successfully brings down the incorrect rate and accounts for retrieval failure.

\begin{figure*}[t]
    \centering
    \includegraphics[width=\linewidth]{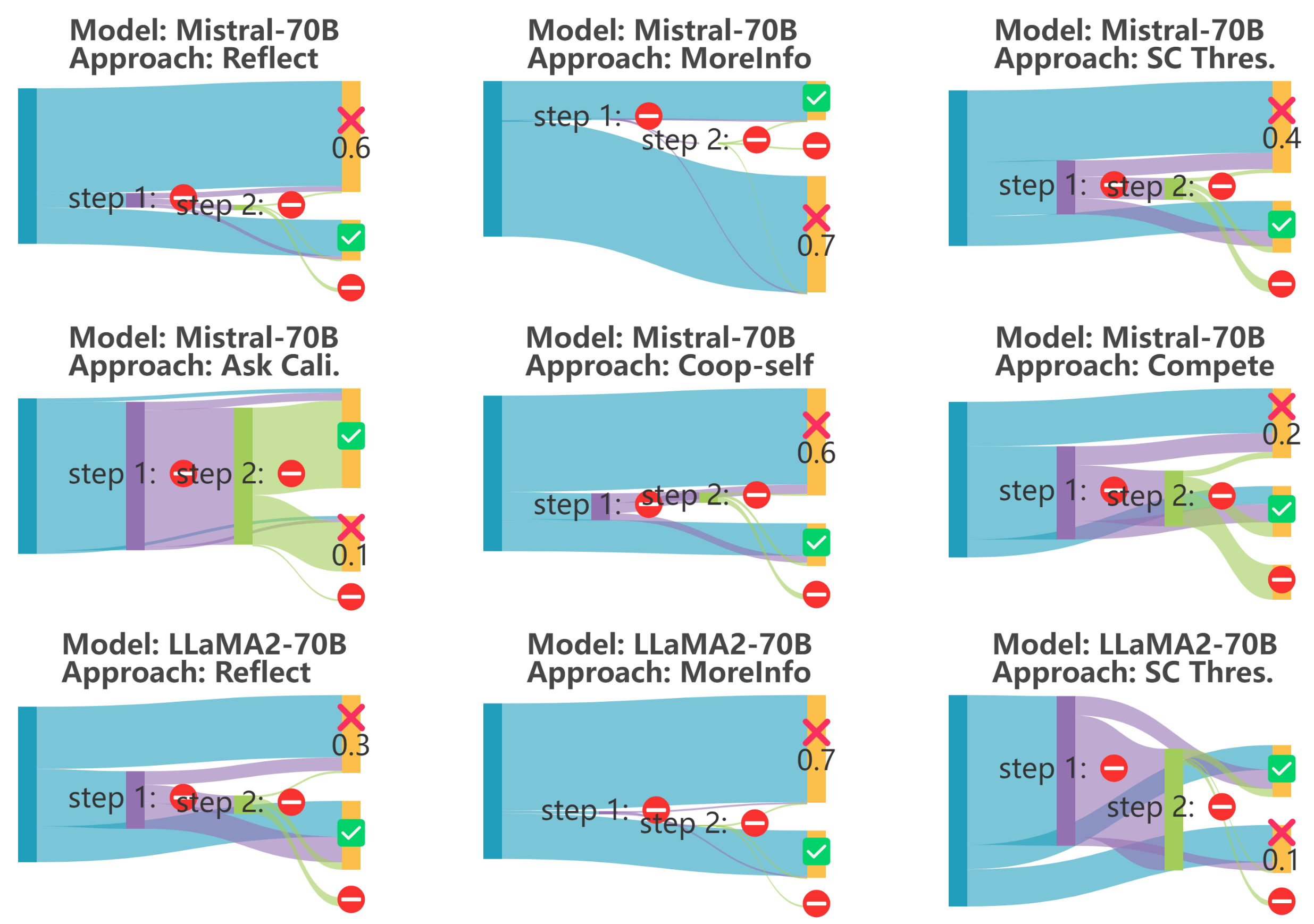}
    \includegraphics[width=\linewidth]{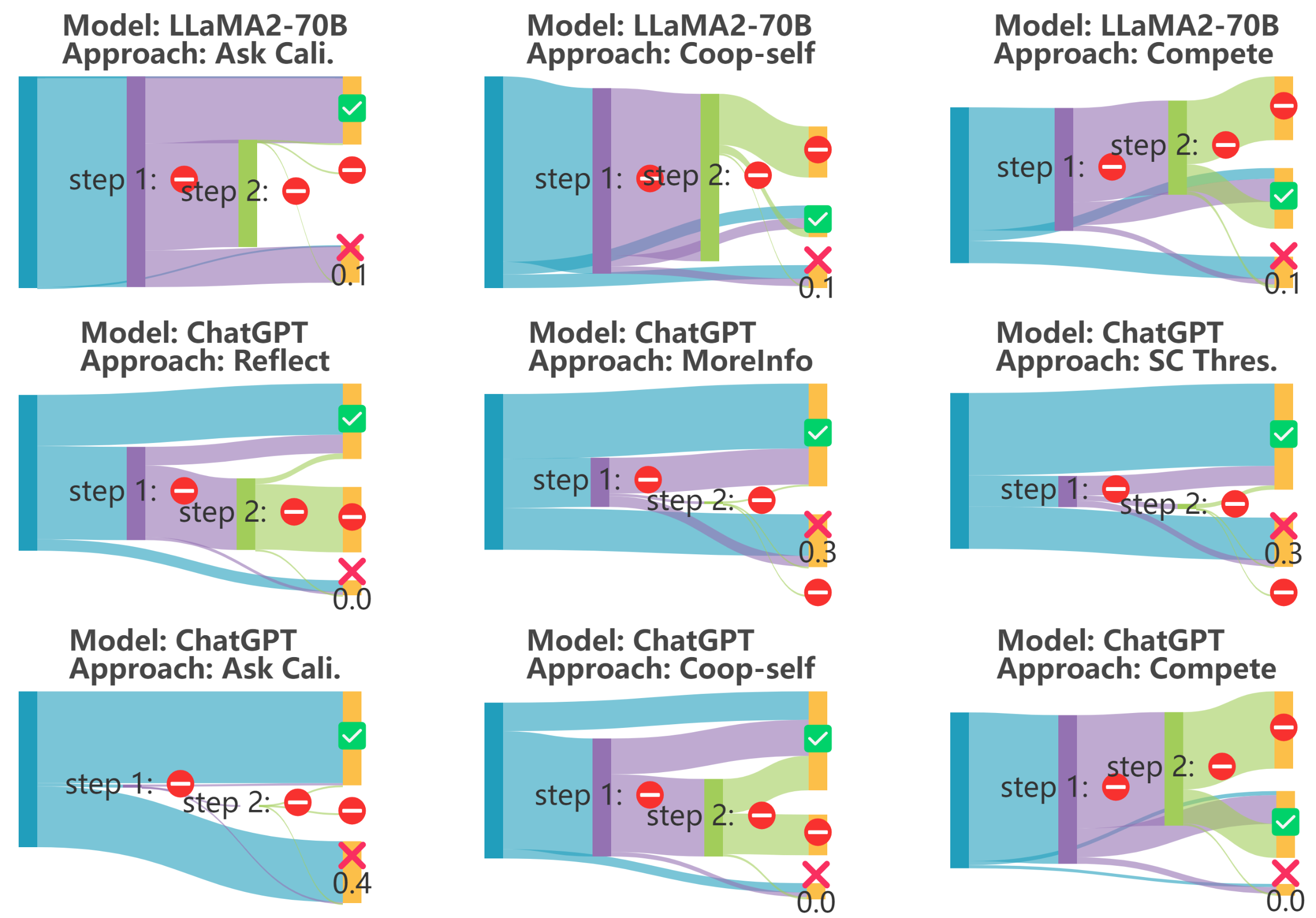}
    \caption{Model performance in the multi-hop knowledge reasoning setting.}
    \label{fig:multi-hop_cont}
\end{figure*}

\paragraph{Abstain and Multi-Hop (cont.)} We present more results with the multi-hop abstention in Figure \ref{fig:multi-hop_cont}. it is demonstrated that our proposed \textsc{Cooperate} and \textsc{Compete} are consistently better at pinpointing knowledge limitations in specific reasoning steps across LLMs and datasets. We further investigate the alignment between model abstain decisions and model answer incorrectness: how many hops did the LLM abstain for and how many hops did the LLM actually answer incorrectly. Table \ref{tab:abstain_incorrect_match} indicates that model abstain decisions match model failures in 73.4\% of the time, showcasing the effectiveness of multi-LLM collaboration for abstention in multi-hop problems.

\begin{table}[]
    \centering
    \resizebox{1\linewidth}{!}{
    \begin{tabular}{lccc}
         \toprule[1.5pt]
         & fail in 1/3 hops & fail in 2/3 hops & fail in 3/3 hops \\ \midrule[0.75pt]
         abstain in 1/3 hops & \textbf{351} & 153 & 12 \\
         abstain in 2/3 hops & 91 & \textbf{512} & 63 \\
         abstain in 3/3 hops & 30 & 20 & \textbf{156} \\ \bottomrule[1.5pt]
    \end{tabular}
    }
    \caption{Alignment between model abstain decisions and incorrect answers regarding the number of hops in multi-hop QA. \textsc{Compete} shows 73.4\% match between abstain decisions and model failures.}
    \label{tab:abstain_incorrect_match}
\end{table}

\paragraph{Held-Out Sets Harms Generalization} In the study we find that most of the baselines would require a held-out set of questions for model tuning, hyperparameter optimization, and more. Specifically, while \textsc{instruction tuning} is one of the strongest baselines in Section \ref{sec:results}, its reliance on a held-out set for training might jeopardize its generalization across knowledge domains and LLMs. To this end, we conduct a generalization study in Figure \ref{fig:domain transfer} where there is a transfer across datasets or LLMs between training and testing. It is illustrated that \textsc{instruction tuning} approaches struggle to generalize across knowledge domains and LLMs, resulting in performance drops sometimes as large as 33.8\% in abstain accuracy. On the contrary, our proposed collaboration-based approaches do not need a held-out set and will not suffer from these generalization limitations.

\paragraph{Minority Opinion} By default, we have $k=3$ feedback passages for each final judge LLM to consider. We find that on the MMLU dataset, the final judge follows the majority opinion in generated feedbacks 84.2\% of the time. We manually examine 10 examples where the judge employed the minority opinion and find that in 8 of the 10 cases, the minority opinion indeed presents new information and should be thus taken into account, showcasing the benefit of having multiple feedbacks through multi-LLM collaboration.

\paragraph{Scaling Laws} We investigate whether LLM abstention with these mechanisms could be improved through scaling, by evaluating three abstain mechanisms with varying sizes of the \textsc{LLaMA2} LLM. Results in Figure \ref{fig:scaling_acc} demonstrate that the abstain performance is not steadily increasing with model size, showcasing that empowering LLMs with abstention abilities is not solely solvable through scaling up.

\begin{figure*}[t]
    \centering
    \includegraphics[width=0.9\linewidth]{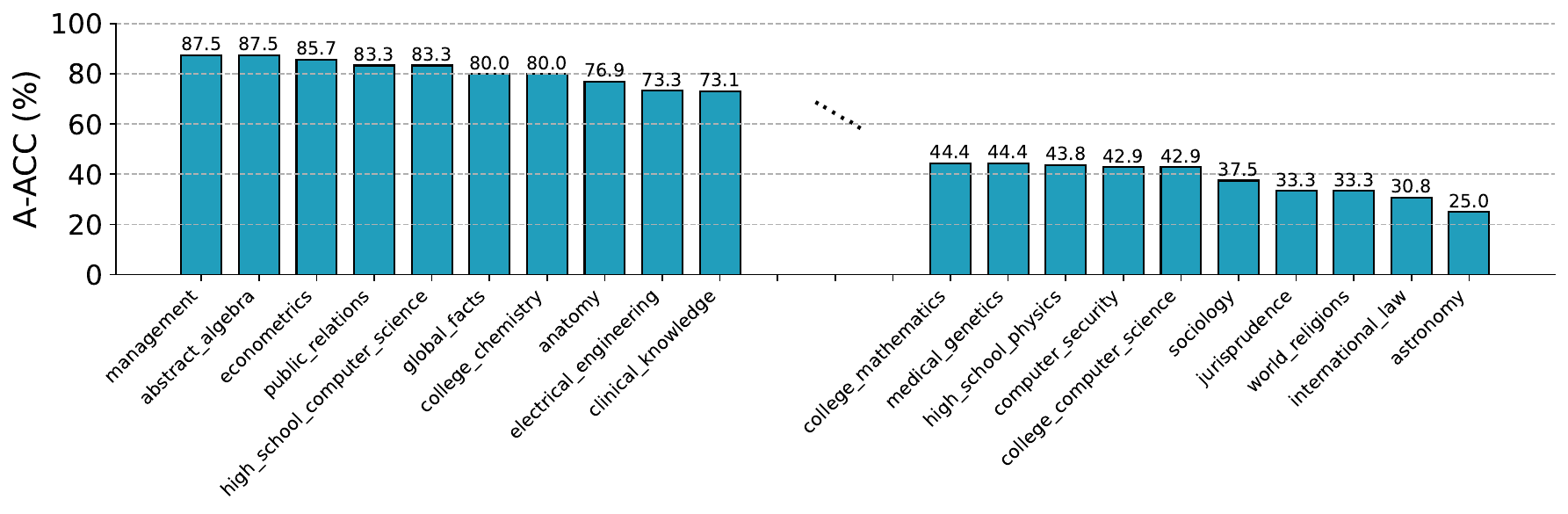}
    \caption{Performance of the \textsc{Compete} mechanism with \emph{Mistral-7B} across the 57 subtasks in MMLU.}
    \label{fig:knowledge_domain}
\end{figure*}

\paragraph{Shift in Knowledge Domains} We investigate whether LLM abstention abilities could be impacted by which knowledge domain the given question is about. Specifically, we present the abstain accuracy metric for different knowledge domains in the MMLU dataset in Figure \ref{fig:knowledge_domain}. It is demonstrated that the abstention decisions are indeed of varying quality given the knowledge domain, from 25\% to 87.5\%, potentially informed by the LLM's underlying knowledge abilities regarding that domain.

\paragraph{Prompt Robustness} Since LLMs could be sensitive to minor changes in prompts \citep{sclar2023quantifying}, we experiment with minor paraphrases of the prompts used in \textsc{Cooperate} and \textsc{Compete} and present the results in Table \ref{tab:prompt_robustness}. It is demonstrated that the proposed abstain mechanisms are relatively robust to minor changes in prompt phrasing.

\begin{table}[]
    \centering
    \resizebox{0.8\linewidth}{!}{
    \begin{tabular}{lcccc}
         \toprule[1.5pt]
         \textbf{Method} & R-Acc & ER & A-Acc & A-F1 \\ \midrule[0.75pt]
         \textsc{Coop-self-1} & .540 & .040 & .544 & .543 \\
         \textsc{Coop-self-2} & .564 & .035 & .498 & .553 \\
         \textsc{Coop-self-3} & .562 & .032 & .504 & .562 \\ \midrule[0.75pt]
         \textsc{Compete-1} & .672 & .088 & .590 & .671 \\
         \textsc{Compete-2} & .698 & .126 & .623 & .680 \\
         \textsc{Compete-3} & .655 & .077 & .577 & .662 \\ \bottomrule[1.5pt]
    \end{tabular}
    }
    \caption{Performance of \textsc{Cooperate} and \textsc{Compete} with variations in the prompt format, on the MMLU dataset with the \emph{Mistral-7B} language model.}
    \label{tab:prompt_robustness}
\end{table}

\paragraph{LLM Overhead} Different abstain mechanisms have different computational overhead and the most expensive part is the number of LLM inference requests for each question. We present a summary in Table \ref{tab:llm_overhead}.

\begin{table}[]
    \centering
    \resizebox{0.8\linewidth}{!}{
    \begin{tabular}{lc|lc}
         \toprule[1.5pt]
         \textbf{Method} & \# & \textbf{Method} & \# \\ \midrule[0.75pt]
         \textsc{NOTA} & 2 & \textsc{reflect} & 2 \\
         \textsc{MoreInfo} & 2 & \textsc{Probs} & 3 \\
         \textsc{SC Thres.} & k+1 & \textsc{Ask Cali.} & 6 \\
         \textsc{Temp.} & 3 & \textsc{Verifier} & 3 \\
         \textsc{Instruct} & 2+\textit{ft} & \textsc{Embedding} & 4 \\
         \textsc{Gen+Match} & 3 & \textsc{Coop-self} & 8 \\
         \textsc{Coop-others} & 2+\textit{o} & \textsc{Compete} & 2+\textit{o} \\ \bottomrule[1.5pt]
    \end{tabular}
    }
    \caption{Overhead of LLM inference requests for different abstain mechanisms. ``\textit{ft}'' denotes additional fine-tuning and ``\textit{o}'' denotes that other reviewer LLMs are also called once.}
    \label{tab:llm_overhead}
\end{table}

\paragraph{Working Examples} We present qualitative examples of \textsc{Cooperate} (Tables \ref{tab:pr-self-1} and \ref{tab:pr-self-2} for \emph{self}, Tables \ref{tab:pr-others-1} and \ref{tab:pr-others-2} for \emph{others}) and \textsc{Compete} (Tables \ref{tab:conflict-1}, \ref{tab:conflict-2}, \ref{tab:conflict-3}, and \ref{tab:conflict-4}).

\section{Experiment Details}
\label{sec:experiment_details}

\subsection{Dataset Details}
\label{subsec:dataset_details}
We employ six datasets in the experiments, four for the main experiments, and two \emph{abstain absolute} datasets.

\begin{itemize}[leftmargin=*]
    \item \textbf{MMLU} \citep{hendrycks2020measuring}: we randomly downsample the official validation and test splits into 1,000 questions each as the held-out set and test set.
    \item \textbf{K-Crosswords} \citep{ding2023knowledge}: we follow the official splits to employ the validation set of 1,094 questions as the held-out set and the test set of 1,007 questions.
    \item \textbf{Hellaswag} \citep{zellers2019hellaswag}: we randomly sample the official validation set into 1,000 questions each as the held-out set and test set.
    \item \textbf{Propaganda} \citep{piskorski2023semeval}: we create a random split of 231 questions as the held-out set and 200 questions as the test set based on the official validation split. We randomly sample three other propaganda tactics accompanied by the correct tactic to form four-way multi-hop choice questions.
    \item \textbf{AmbigQA} \citep{min2020ambigqa}: we randomly downsample the official validation and test splits into 1,000 questions each as the held-out set and test set.
    \item \textbf{ElectionQA23}: we first collect the overview paragraphs of the Wikipedia pages of elections held in 2023\footnote{\href{https://en.wikipedia.org/wiki/List_of_elections_in_2023}{https://en.wikipedia.org/wiki/List\_of\_elections\_in\_2023}} and employ ChatGPT to compose 10 questions for each election in a four-way multiple-choice format. We conduct manual evaluation to remove low-quality or ambiguous questions and create a held-out set of 67 and a test set of 200 questions.
\end{itemize}

\subsection{Model Details}
We employ three large language models to evaluate abstain mechanisms: 1) \emph{Mistral-7B}, through the \textsc{mistralai/Mistral-7b-Instruct-v0.1} checkpoint on Huggingface \citep{wolf2019huggingface}; 2) \emph{LLaMA2-70B}, through the \textsc{meta-llama/LLaMA-2-70b-chat-hf} checkpoint on Huggingface; 3) \emph{ChatGPT}, through the \textsc{gpt-3.5-turbo-instruct} checkpoint on OpenAI API.

\subsection{Implementation Details}
for every single baseline and approach, along with ECE score extraction, prompt for prompting-based and proposed approaches

\begin{itemize}[leftmargin=*]
    \item \textbf{Token Probability}: The token probability of the answer token(s) is employed as $p(\boldsymbol{a})$. The abstain likelihood is obtained through $1-p(\boldsymbol{a})$.
    \item \textbf{Temperature Scaling}: We search for an optimal $\tau$ value from 0.1 to 10 on the held-out set $\mathcal{H}$. The abstain likelihood is obtained through $1-p(\boldsymbol{a})$ after temperature scaling.
    \item \textbf{Ask for Calibration}: We follow \citet{tian2023just} and employ the prompt in Table \ref{tab:prompt_ask_for_calibration} to elicit verbalized confidence scores. The abstain likelihood is obtained through $1-p(\boldsymbol{a})$ with verbalized confidence scores.
    \item \textbf{Hidden Layers}: We employ the feature-extraction pipeline with HuggingFace \citep{wolf2019huggingface} to extract $\mathbf{e}_{\boldsymbol{q}}$. A linear layer of size (\textsc{dim}, 2) is then employed for linear probing where \textsc{dim} is the dimension of $\mathbf{e}_{\boldsymbol{q}}$. The abstain likelihood is obtained as the $\ell = \textit{reject}$ likelihood after softmax.
    \item \textbf{External Verifier}: We employ RoBERTa (\textsc{roberta-base}) \citep{liu2019roberta} as $\mathrm{LLM}-\textit{enc}$ for binary classification. The abstain likelihood is obtained as the $\ell = \textit{reject}$ likelihood with the \textsc{[CLS]} token.
    \item \textbf{Instruction Tuning}: We fine-tune ChatGPT (\textsc{gpt-3.5-turbo-instruct}) with the instruction tuning triplets. The \emph{instruction} is ``Answer the following question. If you don't have enough knowledge, abstain by saying 'sorry, I don't have enough knowledge to answer this question.''' The \emph{abstain message} is ``sorry, I don't have enough knowledge to answer this question.'' The abstain likelihood is obtained as the token probability of the answer token, or that of \emph{``sorry''} if it abstains.
    \item \textbf{Self-Reflect}: Following \citet{kadavath2022language}, we employ the prompt in Table \ref{tab:prompt_self_reflect}. The abstain likelihood is obtained as the token probability of choosing ``B''.
    \item \textbf{More Information}: Following \citet{Feng2023KnowledgeCF}, we employ the prompt in Table \ref{tab:prompt_more_information}. The abstain likelihood is obtained as the token probability of ``yes''.
    \item \textbf{Generate and Match}: We employ the prompt in Table \ref{tab:prompt_generate_and_match}. The abstain likelihood is obtained as the probability of ``no'', or $1-$ that of ``yes''.
    \item \textbf{None-of-the-Above}: The abstain likelihood is obtained as the probability of choosing the NOTA option.
    \item \textbf{Self-Consistency Threshold}: We employ $k=5$ chain-of-thought paths by default. The abstain likelihood is obtained as $1-\mathrm{plu}(\mathrm{LLM},\boldsymbol{q},k)$.
    \item \textbf{Cooperate}: We employ the prompt in Tables \ref{tab:prompt_pr_self} and \ref{tab:prompt_pr_others} for the \emph{self} and \emph{others} modes. The abstain likelihood is obtained as the token probability of choosing ``B''.
    \item \textbf{Compete}: We employ the prompt in Table \ref{tab:prompt_conflict_reflection} and employ three alternative answers by default. The abstain likelihood is obtained as the probability of the new answer if it changes, or $1-$ that of the original answer if it doesn't change, then averaged across all alternative answers.
\end{itemize}

\newpage

\begin{table*}[ht]
\small
\begin{tabularx}{\textwidth}{m{15.6cm}}
\toprule[1pt]

Provide your best guess for the following question. Give ONLY the guess, no other words or explanation. \\ \\

For example: \\ \\

Guess: <most likely guess, as short as possible; not a complete sentence, just the guess!> \\ \\

The question is:  <question> \\ \\

\emph{[LLM-generated answer]} \\ \\

Provide the probability that your guess is correct. Give ONLY the probability, no other words or explanation. \\ \\

For example: \\ \\

Probability: <the probability between 0.0 and 1.0 that your guess is correct, without any extra commentary whatsoever; just the probability!> \\
Probability: \emph{[LLM-generated probability]} \\
\bottomrule[1pt]
\end{tabularx}
\vspace{3pt}
\caption{Prompt for \textsc{Ask for Calibration}.}
\label{tab:prompt_ask_for_calibration}
\end{table*}

\begin{table*}[ht]
\small
\begin{tabularx}{\textwidth}{m{15.6cm}}
\toprule[1pt]
<question> \\ \\

\emph{[LLM-generated answer]} \\
The above answer is: \\
A. True\\
B. False\\
The answer is \emph{[LLM-generated A/B]} \\
\bottomrule[1pt]
\end{tabularx}
\vspace{3pt}
\caption{Prompt for \textsc{Self-Reflect}.}
\label{tab:prompt_self_reflect}
\end{table*}

\begin{table*}[ht]
\small
\begin{tabularx}{\textwidth}{m{15.6cm}}
\toprule[1pt]
<question> \\ \\

Do you need more information to answer this question? (Yes or No)\\
\emph{[LLM-generated yes/no]} \\
\bottomrule[1pt]
\end{tabularx}
\vspace{3pt}
\caption{Prompt for \textsc{More Information}.}
\label{tab:prompt_more_information}
\end{table*}

\begin{table*}[ht]
\small
\begin{tabularx}{\textwidth}{m{15.6cm}}
\toprule[1pt]
<question without multiple-choice options> \\ \\

Proposed answer: \emph{[LLM-generated answer]} \\ \\

<options> \\
Does the proposed answer exist in the options? \\
\emph{[LLM-generated yes/no]} \\
\bottomrule[1pt]
\end{tabularx}
\vspace{3pt}
\caption{Prompt for \textsc{Generate and Match}.}
\label{tab:prompt_generate_and_match}
\end{table*}

\begin{table*}[ht]
\small
\begin{tabularx}{\textwidth}{m{15.6cm}}
\toprule[1pt]
// obtain proposed answer \\ \\

Question: <question> \\
Answer: \emph{[generated proposed answer]} \\ \\

// obtain feedback from self-specialized experts \\ \\

for domain in ["factual information", "commonsense knowledge", "mathematical knowledge"]: \\
\hspace{20pt}        Generate some knowledge about the question, focusing on <domain>: \emph{[generated domain knowledge]} \\ \\

\hspace{20pt}         Knowledge: <generated domain knowledge> \\
\hspace{20pt}         Question: <question> \\
\hspace{20pt}         Answer: <generated proposed answer> \\
\hspace{20pt}         Please review the proposed answer and provide feedback on its correctness. \\
\hspace{20pt}         Feedback: \emph{[generated feedback]} \\ \\

// area-chairing for abstain decision \\ \\

Question: <question> \\
Proposed Answer: <generated proposed answer> \\ \\

Feedback 1: <generated feedback from expert 1> \\
... \\
Feedback k: <generated feedback from expert k> \\ \\

Based on the feedback, the proposed answer is: \\
A. True \\
B. False \\
The answer is \emph{[A/B]}. \\
\bottomrule[1pt]
\end{tabularx}
\vspace{3pt}
\caption{Prompt for \textsc{Cooperate}-\emph{self}.}
\label{tab:prompt_pr_self}
\end{table*}

\begin{table*}[ht]
\small
\begin{tabularx}{\textwidth}{m{15.6cm}}
\toprule[1pt]
// obtain proposed answer \\ \\

Question: <question> \\
Answer: \emph{[generated proposed answer]} \\ \\

// obtain feedback from other LLMs as reviewers

for llm in list\_of\_other\_llms: \\ \\
        
\hspace{20pt}        Question: <question> \\
\hspace{20pt}        Answer: <generated proposed answer> \\
\hspace{20pt}        Please review the proposed answer and provide feedback on its correctness. \\
\hspace{20pt}        Feedback: \emph{[generated feedback using llm]} \\ \\

// area-chairing for abstain decision \\ \\

Question: <question> \\
Proposed Answer: <generated proposed answer> \\ \\

Feedback 1: <generated feedback from llm 1> \\
... \\
Feedback k: <generated feedback from llm k> \\ \\

Based on the feedback, the proposed answer is: \\
A. True \\
B. False \\
The answer is \emph{[A/B]}. \\
\bottomrule[1pt]
\end{tabularx}
\vspace{3pt}
\caption{Prompt for \textsc{Cooperate}-\emph{others}.}
\label{tab:prompt_pr_others}
\end{table*}

\begin{table*}[ht]
\small
\begin{tabularx}{\textwidth}{m{15.6cm}}
\toprule[1pt]
// obtain proposed answer \\ \\

Question: <question> \\
Answer: \emph{[generated proposed answer]} \\ \\

// obtain alternative answers

if multiple-choice: \\
\hspace{20pt}          <alternative answer> = randomly select another unchosen answer \\
else: \\
\hspace{20pt}          Question: <question> \\
\hspace{20pt}          Answer: <generated proposed answer> \\
\hspace{20pt}          Please propose an alternative answer: \emph{[alternative answer]} \\ \\

// obtain a supporting paragraph for the alternative answer \\ \\

Question: <question> \\
Generate a knowledge paragraph about <alternative answer>: \emph{[generated alternative passage]} \\ \\

// challenge! \\ \\

Answer the question with the following knowledge: feel free to ignore irrelevant or wrong information. \\ \\

Knowledge: <generated alternative passage> \\
Question: <question> \\
Answer: \emph{[new generated answer]} \\ \\

// abstain decision \\

if <new generated answer> == <genearted proposed answer>: \\
\hspace{20pt}          abstain = False \\
else: \\
\hspace{20pt}          abstain = True \\
\bottomrule[1pt]
\end{tabularx}
\vspace{3pt}
\caption{Prompt for \textsc{Compete}.}
\label{tab:prompt_conflict_reflection}
\end{table*}

\begin{table*}[ht]
\small
\begin{tabularx}{\textwidth}{m{15.6cm}}
\toprule[1pt]
Question: The following information pertains to Dash Co.'s utility bills: Period covered Amount Date paid April 16 ‚Äì May 15 \$5000 June 1 May 16 ‚Äì June 15 \$6000 July 1 June 16 ‚Äì July 15 \$8000 August 1 What is the amount that Dash should report as a liability in its June 30 balance sheet? \\
A: \$6,000\\
B: \$7,000\\
C: \$10,000\\
D: \$14,000\\
Choose one answer from the above choices. The answer is B: \$7,000\\ \\

Feedback 1: The correct answer is actually A: \$6,000. The amount that Dash should report as a liability in its June 30 balance sheet is the utility bill for the period of May 16 - June 15, which is \$6,000. The subsequent utility bills for June 16 - July 15 and July 16 - August 15 will be reported as liabilities in future balance sheets.\\ \\

Feedback 2: The answer is incorrect. To determine the liability that Dash should report in its June 30 balance sheet, we need to look at the amount that is unpaid as of June 30. According to the information given, the utility bill for April 16 - May 15 was paid on June 1. Therefore, Dash should report the unpaid amount for the period May 16 - June 15 as a liability on its June 30 balance sheet. This amount is \$6,000. Thus, the correct answer is A: \$6,000.\\ \\

Feedback 3: The proposed answer is incorrect.\\ \\

Based on the feedback, the proposed answer is:\\
A. True\\
B. False\\
The answer is\\
 B. False\\
\bottomrule[1pt]
\end{tabularx}
\vspace{3pt}
\caption{Example one for \textsc{Cooperate}-self. The reviewers provided different feedback, while the area chair LLM merged into a final abstain decision.}
\label{tab:pr-self-1}
\end{table*}

\begin{table*}[ht]
\small
\begin{tabularx}{\textwidth}{m{15.6cm}}
\toprule[1pt]
Question: Compared with the current in an LED lamp the current in the connecting wire is \\
A: less.\\
B: more.\\
C: the same.\\
D: Need more information.\\
Choose one answer from the above choices. The answer is A: less. \\ \\

Feedback 1: The proposed answer, A: less, is incorrect. The answer C: the same, would be more appropriate. In an LED lamp, the current flows from the power source through the connecting wire and into the LED. The current in the wire would be the same as the current in the LED if there are no resistive elements in the wire causing a drop in voltage.\\ \\

Feedback 2: The proposed answer is incorrect. The statement does not provide enough information to determine the current in the connecting wire compared to the current in an LED lamp. Therefore, the correct answer is D: Need more information.\\ \\

Feedback 3: The proposed answer is correct. In an LED lamp, the current in the connecting wire is typically less compared to the current in the LED itself. This is because the LED lamp is designed to limit the amount of current flowing through the LED to ensure proper operation and prevent damage. The connecting wire acts as a conductor to allow the flow of current from the power source to the LED, but it does not have as high a current as the LED itself.\\ \\

Based on the feedback, the proposed answer is:\\
A. True\\
B. False\\
The answer is\\
 B. False.\\
\bottomrule[1pt]
\end{tabularx}
\vspace{3pt}
\caption{Example two for \textsc{Cooperate}-self. The reviewers provided different feedback, while the area chair LLM merged into a final abstain decision.}
\label{tab:pr-self-2}
\end{table*}

\begin{table*}[ht]
\small
\begin{tabularx}{\textwidth}{m{15.6cm}}
\toprule[1pt]
Question: The following information pertains to Dash Co.'s utility bills: Period covered Amount Date paid April 16 ‚Äì May 15 \$5000 June 1 May 16 ‚Äì June 15 \$6000 July 1 June 16 ‚Äì July 15 \$8000 August 1 What is the amount that Dash should report as a liability in its June 30 balance sheet? \\
A: \$6,000\\
B: \$7,000\\
C: \$10,000\\
D: \$14,000\\
Choose one answer from the above choices. The answer is B: \$7,000\\ \\

Feedback 1: The answer provided is not correct. The correct amount that Dash should report as a liability in its June 30 balance sheet is \$6,000. This is because the amount for the period of May 16 - June 15 has already been paid on June 1, and does not need to be reported as a liability. The only amount that is still outstanding and needs to be reported is the amount for the period of June 16 - July 15, which is \$6,000. Therefore, the correct answer is A: \$6,000.\\ \\

Feedback 2: The proposed answer is incorrect. Dash should report \$7,000 as a liability in its June 30 balance sheet, but not \$6,000.\\ \\

Here's the reasoning:\\ \\

1. The period covered by the utility bills is May 16 ‚Äì June 15.\\
2. The amount due is \$6,000, and Dash paid \$5,000 on May 16 ‚Äì May 15 and \$1,000 on June 16 ‚Äì June 15.\\
3. The date paid for the \$6,000 bill is June 16 ‚Äì June 15 and the date paid for the \$1,000 bill is June 16 ‚Äì June 15.\\
4. Since the payment of the \$1,000 bill has already\\ \\

Feedback 3: The correct answer is indeed B: \$7,000.\\
The explanation is as follows:\\ \\

The June 30 balance sheet should report a liability for the utility bills that have been incurred but not yet\\ \\

Based on the feedback, the proposed answer is:\\
A. True\\
B. False\\
The answer is\\
 B. False.\\
\bottomrule[1pt]
\end{tabularx}
\vspace{3pt}
\caption{Example one for \textsc{Cooperate}-others. The reviewers provided different feedback, while the area chair LLM merged into a final abstain decision.}
\label{tab:pr-others-1}
\end{table*}

\begin{table*}[ht]
\small
\begin{tabularx}{\textwidth}{m{15.6cm}}
\toprule[1pt]
Question: Who kills Tony at the end of the film 'West Side Story'? \\
A: Riff\\
B: Chino\\
C: Bernardo\\
D: He kills himself\\
Choose one answer from the above choices. The answer is B: Chino\\ \\

Feedback 1: The proposed answer is correct. Chino is the one who kills Tony at the end of the film 'West Side Story'.\\ \\

Feedback 2: The answer provided is incorrect. In the film 'West Side Story', it is Bernardo who kills Tony. Thank you for providing the correct answer.\\ \\

Feedback 3: The proposed answer is incorrect. Tony is killed by Bernardo, not Chino. The correct answer is C: Bernardo.\\ \\

New proposed answer:\\
Answer: Who kills Tony at the end of the film 'West Side Story'\\ \\

Based on the feedback, the proposed answer is:\\
A. True\\
B. False\\
The answer is\\
 B. False. The correct answer is C: Bernardo.\\
\bottomrule[1pt]
\end{tabularx}
\vspace{3pt}
\caption{Example two for \textsc{Cooperate}-others. The reviewers provided different feedback, while the area chair LLM merged into a final abstain decision.}
\label{tab:pr-others-2}
\end{table*}

\begin{table*}[ht]
\small
\begin{tabularx}{\textwidth}{m{15.6cm}}
\toprule[1pt]
Question: The following information pertains to Dash Co.'s utility bills: Period covered Amount Date paid April 16 ‚Äì May 15 \$5000 June 1 May 16 ‚Äì June 15 \$6000 July 1 June 16 ‚Äì July 15 \$8000 August 1 What is the amount that Dash should report as a liability in its June 30 balance sheet? \\
A: \$6,000\\
B: \$7,000\\
C: \$10,000\\
D: \$14,000\\
Generate a knowledge paragraph about A.\\
Dash Co. should report \$6,000 as a liability in its June 30 balance sheet. This is because the utility bills for the period from May 16 to June 15 amount to \$6,000. The balance sheet is a financial statement that provides an overview of a company's financial position at a specific point in time. Liabilities are obligations or debts owed by a company, which are classified as current liabilities if they are expected to be settled within one year. In this case, the utility bill for the period ending on June 15 falls within the one-year timeframe and should therefore be reported as a current liability. Therefore, the correct answer is A: \$6,000.\\
New answer:
 A: \$6,000.\\
\bottomrule[1pt]
\end{tabularx}
\vspace{3pt}
\caption{Example one for \textsc{Compete}. The LLM is impacted by the alternative answer and should abstain.}
\label{tab:conflict-1}
\end{table*}

\begin{table*}[ht]
\small
\begin{tabularx}{\textwidth}{m{15.6cm}}
\toprule[1pt]
Question: A flashing red traffic light signifies that a driver should do what?\\
A: stop\\
B: speed up\\
C: proceed with caution\\
D: honk the horn\\
Generate a knowledge paragraph about D.\\
D: Honking the horn is the appropriate action when encountering a flashing red traffic light. This is an important traffic rule that drivers must follow to maintain safety on the roads. When encountering a flashing red light, it is crucial for all drivers to be aware of traffic signals and adhere to their instructions to ensure a smooth flow of traffic and prevent any potential hazards. Honking the horn, on the other hand, is generally used as a way to alert other drivers or pedestrians of potential dangers or to communicate with them in certain situations.\\
New answer:
A: stop.\\
\bottomrule[1pt]
\end{tabularx}
\vspace{3pt}
\caption{Example two for \textsc{Compete}. The LLM is not impacted by the alternative answer.}
\label{tab:conflict-2}
\end{table*}

\begin{table*}[ht]
\small
\begin{tabularx}{\textwidth}{m{15.6cm}}
\toprule[1pt]
Question: Kelsen’s theory of law is called pure theory because Kelsen: \\
A: Purely discussed jurisprudence only\\
B: Defined law in according with morality and purity\\
C: Separated law from religion, ethics, sociology and history\\
D: Discussed law purely in terms of justice\\
Generate a knowledge paragraph about D.\\
Kelsen's theory of law is called pure theory because he discussed law purely in terms of justice. According to Kelsen, law should be analyzed and understood independently of any moral or ethical considerations. He believed that law should be studied and interpreted solely through its internal system of norms, without any reference to external moral values or social contexts.\\ \\

Kelsen's pure theory of law aimed to provide a "science of law" by separating law from other disciplines such as religion, ethics, sociology, and history. He argued that law should be treated as a self-contained system, with its own hierarchy of norms and principles. This meant that the validity of laws should be determined based on their consistency with higher-level norms, rather than any inherent moral or ethical qualities.\\ \\

By focusing purely on the concept of justice, Kelsen attempted to provide an objective and neutral framework for understanding and analyzing legal systems. He believed that by divorcing law from subjective moral judgments, his theory would provide a more rigorous and scientific approach to the \\
New answer:
D: Discussed law purely in terms of justice. \\
\bottomrule[1pt]
\end{tabularx}
\vspace{3pt}
\caption{Example three for \textsc{Compete}. The LLM is impacted by the alternative answer and should abstain.}
\label{tab:conflict-3}
\end{table*}

\begin{table*}[ht]
\small
\begin{tabularx}{\textwidth}{m{15.6cm}}
\toprule[1pt]
Question: Suppose you live on the Moon. How long is a day (i.e. from sunrise to sunrise)? \\
A: about 18 years\\
B: 24 hours\\
C: 29 Earth days\\
D: a year\\
Generate a knowledge paragraph about A.\\
Option A: about 18 years\\ \\

If you were to live on the Moon, a day would last about 18 years. This is due to a phenomenon called tidal locking. Tidal locking occurs when the gravitational force between two objects, such as the Earth and the Moon, causes one object to always face the other with the same side. In the case of the Moon, it is tidally locked to the Earth, meaning one side is constantly facing our planet.\\ \\

This tidal locking also means that the Moon's rotation period matches its orbital period around the Earth, taking approximately 27.3 Earth days to complete one full revolution. However, despite this prolonged rotation period, it is still necessary to define a day from a sunrise to a sunrise, just as we do on Earth.\\ \\

Since the Moon does not have an atmosphere to scatter sunlight and bring about the distinct sunrise and sunset we are accustomed to, the definition of a lunar day can be somewhat arbitrary. One way to measure it is based on\\
New answer:
 B: 24 hours.\\
\bottomrule[1pt]
\end{tabularx}
\vspace{3pt}
\caption{Example four for \textsc{Compete}. The LLM is not impacted by the alternative answer.}
\label{tab:conflict-4}
\end{table*}

\end{document}